%% file: main.tex
\documentclass{article}

\usepackage{amsthm}
\usepackage{graphicx}
\usepackage{subcaption}
\usepackage{amssymb}
\usepackage{bbm}
\usepackage{url}
\usepackage{amsmath}
\usepackage{booktabs}
\usepackage{mathtools}
\usepackage{hyperref}

\DeclareMathOperator*{\argmax}{arg\,max}
\DeclareMathOperator*{\argmin}{arg\,min}

\newtheorem{definition}{Definition}
\newtheorem{lemma}{Lemma}
\newtheorem{theorem}{Theorem}
\newtheorem{example}{Example}

\newcommand{\lif}[0]{
  \leftarrow
}

\newcommand{\rif}[0]{
  \rightarrow
}
\newcommand{\vt}[1]{
\mathbf{#1}
}
\newcommand{\fn}[1]{
{\it #1}
}

\newcommand{\xor}[0]{
  \oplus
}
\newcommand{\fzand}[0]{
  \wedge
}
\newcommand{\bAnd}[0]{
  \bigwedge
}
\newcommand{\fzor}[0]{
  \vee
}
\newcommand{\bOr}[0]{
  \bigvee
}

\newcommand{\fziff}[0]{
  \leftrightarrow
}

\newcommand{\pr}[1]{
  \mathrm{#1}
}
\newcommand{\imgobj}[1]{\includegraphics[width=0.35cm]{#1}}

\newcommand{\En}[0]{ 
  \fn{E}
}

\newcommand{\dtime}[1]{\textcolor{blue}{}}

\newcommand{\xdownarrow}[1]{%
  {\left\downarrow\vbox to #1{}\right.\kern-\nulldelimiterspace}
}
\usepackage{authblk}
\title{Reasoning in Neurosymbolic AI}

\author[1]{Son Tran}
\author[2]{Edjard Mota}
\author[1]{Artur d'Avila Garcez}

\affil[1]{School of Information Technology, Deakin University, Victoria, 3125, Melbourne, Australia}
\affil[2]{Instituto de Computação, Universidade Federal do Amazonas, 69067-005, Manaus, Brazil}
\affil[3]{Department of Computer Science, City St George's, University of London, EC1V 0HB, London, UK}

\begin{document}

\maketitle

\begin{abstract}
Knowledge representation and reasoning in neural networks have been a long-standing endeavor which has attracted much attention recently. The principled integration of reasoning and learning in neural networks is a main objective of the area of neurosymbolic Artificial Intelligence (AI). In this chapter, a simple energy-based neurosymbolic AI system is described that can represent and reason formally about any propositional logic formula. This creates a powerful combination of learning from data and knowledge and logical reasoning. We start by positioning neurosymbolic AI in the context of the current AI landscape that is unsurprisingly dominated by Large Language Models (LLMs). We identify important challenges of data efficiency, fairness and safety of LLMs that might be addressed by neurosymbolic reasoning systems with formal reasoning capabilities. We then discuss the representation of logic by the specific energy-based system, including illustrative examples and empirical evaluation of the correspondence between logical reasoning and energy minimization using Restricted Boltzmann Machines (RBM). The system, called Logical Boltzmann Machine (LBM), can find all satisfying assignments of a class of logical formulae by searching through a very small percentage of the possible truth-value assignments. Learning from data and knowledge in LBM is also evaluated empirically and compared with a purely-symbolic, a purely-neural and a state-of-the-art neurosymbolic system, achieving better learning performance in five out of seven data sets. Results reported in this chapter in an accessible way are expected to reignite the research on the use of neural networks as massively-parallel models for logical reasoning and to promote the principled integration of reasoning and learning in deep networks. LBM is also evaluated in the role of an interpretable neural \textit{module} that can be added on top of complex neural networks such as convolutional networks and encoder-decoder networks to implement any given set of logical constraints e.g. fairness or safety requirements. LBM is further evaluated when deployed in the solution of the connectionist Boolean satisfiability (SAT) problem, maximum satisfiability (MaxSAT) and approximate optimization problems when certain logical rules may be given a higher priority or a penalty according to a \textit{confidence value}. We conclude the chapter with a discussion of the importance of positioning neurosymbolic AI within a broader framework of formal reasoning and accountability in AI, discussing the challenges for neurosynbolic AI to tackle the various known problems of reliability of deep learning. We close with an opinion on the risks of AI and future opportunities for neurosymbolic AI. Keywords: Neurosymbolic AI, Restricted Boltzmann Machines, Logical Reasoning, SAT solving, MaxSAT, Energy-based Learning, Constrained Optimization, Modular Deep Learning.
\end{abstract}
\sloppy
\section{What is Reasoning in Neural Networks?}
\input{artur1}

\section{Background: Logic and Restricted Boltzmann Machines}
\input{edjard1}

\section{Symbolic Reasoning with Energy-based Neural Networks}
\input{AAAI}
\section{Logical Boltzmann Machines for MaxSAT}
\input{edjard2}

\section{Integrating Learning and Reasoning in Logical Boltzmann Machines}
\input{firstorder}

\section{Challenges for Neurosymbolic AI} 
\input{artur2}

\section{Conclusion}
\input{artur3}

\bibliographystyle{plain}
\bibliography{biblio,kr-sample}

\end{document}

%% file: artur1.tex
Increasing attention has been devoted in recent years to knowledge representation and reasoning in neural networks. The principled integration of reasoning and learning in neural networks is a main objective of the field of neurosymbolic Artificial Intelligence (AI) \cite{Garcez_2008, OdenseAI}. In neurosymbolic AI, typically, an algorithm is provided that translates some form of symbolic knowledge representation into the architecture and initial set of parameters of a neural network. Ideally, a theorem then shows that the neural network can be used as a massively-parallel model of computation capable of reasoning about such knowledge. Finally, when trained with data and knowledge, the network is expected to produce better performance, either a higher accuracy or faster learning than when trained from data alone. Symbolic knowledge may be provided to a neural network in the form of general rules which are known to be true in a given domain, or rules which are expected to be true across domains when performing transfer and continual learning. When rules are not available to start with, they can be extracted from a trained network. When rules are contradicted by data, they can be revised as part of the learning process. This has been shown to offer a flexible framework whereby knowledge and data, neural networks and symbolic descriptions are combined, leading to a better understanding of complex network models with the interplay between learning and reasoning. 

This chapter includes a general discussion of how neurosymbolic AI can contribute to the goals of reasoning in neural networks and a specific illustration of a neurosymbolic system for reasoning in propositional logic with restricted Boltzmann machines (RBMs) \cite{Smolensky_1995}. We will describe a neurosymbolic system, called Logical Boltzmann Machines (LBM), capable of (i) representing any propositional logic formula into a restricted Boltzmann machine, (ii) reasoning efficiently from such formula, and (iii) learning from such knowledge representation and data. LBM comes with an algorithm to translate any set of propositional logical formulae into a Boltzmann machine and a proof of equivalence between the logical formulae and the energy-based connectionist model; in other words, a proof of soundness of the translation algorithm from logical formulae to neural networks. Specifically, the network is shown to assign minimum energy to the assignments of truth-values that satisfy the formulae. This provides a new way of performing reasoning in symmetrical neural networks by employing the network to search for the models of a logical theory, that is, to search for the assignments of truth-values that map the logical formulae to $true$.\footnote{We use the term \emph{model} to refer to logical models and to neural network models. When the intended meaning is not clear from the context, we shall use the term \emph{logical model}.} If the number of variable is small, inference can be carried out analytically by sorting the free-energy of all possible truth-value assignments. Otherwise, Gibbs sampling is applied in the search for logical models. We start, however, with a general discussion of reasoning in current AI including large language models. 

\subsection{Reasoning in Large Language Models}

Since the release of GPT4 by OpenAI in March 2023, a fierce debate developed around the risks of AI, Big Tech companies released various proprietary and open-source competitors to ChatGPT, and the European Union passed the regulatory AI Act in record time. Leading figures disagreed on what should be done about the risks of AI. Some claimed that Big Tech is best placed to take care of safety, others argued in favor of open source, and others still argued for regulation of AI and social media. As society contemplates the impact of AI on everyday life, the secrecy surrounding AI technology fueled fears of existential risk and even claims of an upcoming AI bubble burst. Large Language Models (LLMs) such as ChatGPT, Gemini, Claude, Mistral and DeepSeek are a great engineering achievement, are impressive at text summarization and language translation, may improve productivity of those who are knowledgeable enough to spot the LLM’s mistakes, but have great potential to deceive those who aren’t. 

There are various technical and non-technical reasons why LLMs and current AI may not be deployed in practice: lack of trust or fairness, reliability issues and public safety as in the case of self-driving cars that use the same technology as LLMs. Fixing reliability issues case-by-case with Reinforcement Learning has proved to be too costly. A common risk mitigation strategy has been to adopt a \textit{human-in-the-loop} approach: making sure that a human is ultimately responsible for decision making. However, in the age of Agentic AI, where at least some decisions are made by the machine, simply apportioning blame or liability to a human does not address the problem. It is necessary to empower the user of AI, the data scientist and the domain expert to be able to interpret, question and if necessary intervene in the AI system. Neural networks that are accompanied by symbolic descriptions and sound reasoning capabilities will be an important tool in this process of empowering users of AI.

Consider LLMs’ ability to produce code. If GPT4 was allowed to work, not as a stand-alone computer program, but in a loop whereby the code can be executed and data collected from execution to improve the code automatically, one can see how such self-improving LLM with autonomy may pose a serious risk to current computer systems. Recent experiments, however, indicated that the opposite, self-impairing, may also happen in practice, producing a degradation in performance. We will argue that the emerging field of neurosymbolic AI can address such failures and that there must be a better way, other than very costly post-hoc model alignment, of achieving AI that can offer certain logical guarantees to network training. 

LLMs have been considered to be general purpose because they will provide an answer to any question. They do that by doing only one thing: predicting the probability of the next word (token) in a sentence. Having made a choice of the next word, LLMs will apply the same calculations recursively to build larger sentences. They are called auto-regressive machine learning models because they perform regression on the discrete tokens to learn such probabilities, and apply recursively the learned function $f$ to choose the word that comes at time t+1 given the words that are available at time t, that is, $x_{t+1}=f(x_t)$. Artificial General Intelligence (AGI), however, is best measured by the ability to adapt to novelty. It will require effective learning from fewer data, the ability to reason reliably about the knowledge that has been learned, the extraction of compact descriptions from trained networks and the consolidation of knowledge learned from multiple tasks, using analogy to enable extrapolation to new situations at an adequate level of abstraction. It has been almost two years since GPT4 was released. The competition has caught up. Reliable data seem to have been exhausted. Performance increments obtained with increase in scale have not produced AGI. It is fair to say that the ``scale is all you need" claim has not been confirmed. Notwithstanding, domain-specific AI systems that can exhibit intelligence at the level of humans or higher already exist. These systems exhibit intelligence in specialized tasks: targeted medical diagnoses, protein folding, various closed-world two-player strategy games. 

When LLMs make stuff up such as non-existing citations, they are said to hallucinate. AGI will require systems that never hallucinate (that is, reason reliably), that can form long term plans and act on those plans to achieve a goal, and that can handle exceptions as they materialize, addressing shifts in data distribution not case-by-case, but requiring far less data labeling. This is very different from current LLMs that seem to have difficulty handling exceptions. For this reason, hallucinations are not going away and the cost of post-hoc model alignment has spiraled in the last two years.

As a case in point, take the o1 LLM system released by OpenAI in September 2024; o1 was claimed to ``think before it answers” and to be capable of ``truly general reasoning”. Widely seen as a re-branding of the much anticipated GPT5, which was promised to be at AGI level, the little that we know about o1 is that it improved on reasoning and code generation benchmarks, and yet it can be stubbornly poor at simple tasks such as multiplication, formal reasoning, planning or the formidable ARC AGI challenge (see https://arcprize.org/). Let's assume that OpenAI’s o1 system is best described as ``GPT-Go", a pre-trained transformer to which a tree search is incorporated in the style of Google DeepMind’s earlier Alpha-Go system. The tree search uses ``Chain of Thought" (CoT) prompting: generation of synthetic data using the transformer neural network itself in a chain that breaks down a prompt into sub-prompts (sub-problems to be solved in stages). o1's ``thinking” time is presumably needed to build the tree for the CoT. And it's this breaking of the problem into sub-problems that is expected to improve performance on reasoning tasks since this is how reasoning tasks are solved.

Leaving aside the practical question of how long users will be happy to wait for an answer, the main issue with o1 and successors is a lack of reliability of the synthetic data generation and combinatorial nature of CoT: CoT may solve one reasoning task well today only to fail at an analogous reasoning task tomorrow due to simple naming variations \cite{mirzadeh2024gsmsymbolicunderstandinglimitationsmathematical}. With synthetic data generation from GPT-like auto-regressive models having been shown to impair model performance, the quality of the data decreases and the model continues to hallucinate \cite{Shumailov2023TheCO}.

What we are seeing in practice is that eliminating hallucinations is very difficult. And there is another concern: regurgitation. The New York Times (NYT) lawsuit against OpenAI argues that ChatGPT can basically reproduce (regurgitate) copyrighted NYT texts with minimal prompting. Whether regurgitation can be fixed remains to be seen. Efforts in this direction have been focused on a simple technique called RAG (Retrieval Augmented Generation) that fetches facts from external sources. What is clear is that further research is needed to make sense of how LLMs generalize to new situations, to find out whether performance depends on task familiarity or true generalization. In the meantime, there will be many relevant but domain-specific applications of LLMs in areas where the system has been deemed to have been controlled reasonably well or where controlling it isn't crucial. 

In neurosymbolic AI, instead of adjusting the input to fix a misbehaving LLM as done with CoT, the idea is to control the architecture or the loss function of the system. Neurosymbolic AI integrates learning and reasoning to make model development parsimonious by following this recipe: (1) extract symbolic descriptions as learning progresses, (2) reason formally about what has been learned, (3) compress the neural network as knowledge is instilled back into the network. Reasoning in neurosymbolic AI follows the tradition of knowledge representation in AI. It requires the definition of a semantics for deep learning and it measures the capabilities of neural networks w.r.t. formally-defined, sound and approximate reasoning, providing a much needed measure of the accumulation of errors in the AI system.  

\subsection{AI from a Neurosymbolic Perspective}

It is paradoxical that computers have been invented to provide fast calculations and sound reasoning, and yet the latest AI may fail at calculations as simple as multiplication (even though a typical artificial neural network will rely on millions of correct multiplications as part of its internal computations). The first wave of AI in the 1980s was knowledge-based, well-founded and inefficient if compared with deep learning. The second wave from the 2010s was data-driven, distributed and efficient but unsound if compared with knowledge-bases. It is clear that neural networks are here to stay, but the problems with deep learning have been stubbornly difficult to fix using neural networks alone. Next, we discuss how solving these problems will require the use of symbolic AI alongside neural networks. The third wave of AI, we argue, will be neurosymbolic \cite{garcez2020neurosymbolic}.

In order to understand the achievements and limitations of AI, it is helpful to consider the \textit{AGI debate}\footnote{\url{https://www.youtube.com/watch?v=JGiLz_Jx9uI}.} with its focus on what is missing from current AI systems, i.e. the technological innovation that may bring about better AI or AGI. Simply put, such innovation may be described as the ability to apply knowledge learned from a task by a neural network to a novel task without requiring too much data.

As AI experts John Hopfield and Geoff Hinton are awarded the 2024 Nobel Prize for Physics, and AI expert Demis Hassabis is awarded the 2024 Nobel Prize for Chemistry (with David Baker and John Jumper), one can say that the era of computation as the language of science has began. Hassabis led the team at Google DeepMind that created AlphaFold, an AI model capable of predicting with high accuracy the 3D structure of proteins given their amino acid sequence. AlphaFold is arguably the greatest achievement of AI to date, even though it is squarely an application specific (or narrow) AI by comparison with LLMs. From particle physics to drug discovery, energy efficiency and novel materials, AI is being adopted as the process by which scientific research is carried out. However, as noted above, the lack of a description or explanation capable of conveying a deeper sense of understanding of the solution being offered by AI is something that is very unsatisfactory. Computer scientists in a great feat of engineering will solve to a high degree of accuracy very challenging problems in science without necessarily improving their own understanding of the solutions provided by very large neural networks trained on vast amounts of data that are not humanly possible to inspect.

The risks of current AI together with this unsatisfactory lack of explainability confirm the need for neurosymbolic AI as an alternative approach. As mentioned, neurosymbolic AI uses the technology of knowledge extraction to interpret, ask \textit{what-if} questions and if necessary intervene in the AI system, controlling learning in ways that can offer correctness or fairness guarantees and, with this process, producing a more compact, data efficient system. We start to see a shift towards such explainable neurosymbolic AI systems being deployed as part of a risk-based approach. As argued in \cite{AccountAI}, effective regulation goes hand in hand with accountability in AI, the definition of a risk mitigation strategy and the use of technology itself such as explainable AI technology \cite{NgunXAI} to mitigate risks. We shall return to this discussion at the end of the paper.

For more than 20 years, a small group of researchers have been advocating for neurosymbolic AI. Already around the turn of the 21st century, the importance of artificial neural networks as an efficient computational model for learning was clear to that group. But the value of symbol manipulation and abstract reasoning offered by symbolic logic was also obvious to them. Many before them have contributed to neurosymbolic AI. In fact, it could be argued that neurosymbolic AI starts together with connectionism itself, with the aptly titled 1943 paper by McCulloch and Pitts, \textit{A Logical Calculus of the Ideas Immanent in Nervous Activity}, and with John Von Neumann’s 1952 \textit{Lectures on Probabilistic Logics and the Synthesis of Reliable Organisms from Unreliable Components}, indicating that the gap between distributed vector representations (embeddings) and localist symbolic representations in logic was not as big as some might imagine. Even Alan Turing’s 1948 \textit{Intelligent Machinery} introduced a type of neural network called a B-type machine. All of this, of course, before the term Artificial Intelligence was coined ahead of the now famous Dartmouth Workshop in 1956. Since then the field has separated into two: symbolic AI and connectionist AI (or neural networks). This has slowed progress as the two research communities went their separate ways with different conferences, journals and associations. Following the temporary success of symbolic AI in the 1980’s and the success of deep learning since 2015 with its now obvious limitations, the time is right for revisiting the approaches of the founding fathers of computer science and developing neurosymbolic AI that is fit for the 21st century. As a step in this direction, in what follows, we illustrate how a single bi-directional network layer in the form of a restricted Boltzmann machine can implement the full semantics of propositional logic, formally defined.

%% file: edjard1.tex
Differently from general-purpose Large Language Models, domain specific Artificial Intelligence, such as the protein folding AlphaFold system, aims to develop systems for specific purposes, enabling human abilities to handle tasks that might otherwise take many years to solve. This goal of domain specific AI is analogous to the invention of the Archimedean lever, which enhanced physical strength capabilities and has enabled humanity to make leaps in construction, mobility and physical labor. AI can be a mental lever that enhances our ability to deal with problems requiring mental activity in volume or intensity that is difficult to accomplish in feasible time or with precision. Modeling such abstract human mental activity is a highly complex task and we shall focus on representing two well-studied aspects: learning and reasoning. 

A key step in this endeavor is to choose an appropriate language to represent the problem at hand. In the context of this paper, such a choice will be deemed to be suitable if it allows the development of efficient algorithms to perform learning from data and reasoning about what was learned or if it allows one to identify patterns of solutions that will lead to adequate decisions. Traditional AI has separated the study of reasoning and learning with a focus on either knowledge elicitation by hand for the purpose of sound reasoning or statistical learning from large amounts of data. In neurosymbolic AI this artificial separation is removed. The neurosymbolic cycle seeks to enable AI systems to \textit{learn a little} and \textit{reason a little} in integrated fashion. Learning takes place in the usual way within a neural network but reasoning has to be formalized, whether taking place inside or outside the network. Instead of simply measuring \textit{reasoning capabilities} of the networks using benchmarks, neurosymbolic AI networks seek to offer reasoning guarantees of correctness. It is crucial to pay attention to the many years of research in knowledge representation and reasoning within Computer Science logic. While learning may benefit from the use of natural language and other available multimodal data, sound reasoning requires a formal language. A choice of language adequate to the problem influences the system's ability to find a solution.  

Formal logic, particularly Propositional Logic, is the most straightforward language for representing propositions about the problem domain.
Propositional logic is the simplest formal language for representation, a branch of mathematics and logic that deals with simple declarative statements, called propositions, which can be \textit{true} or \textit{false}. As we shall see, in the context of neurosymbolic systems, statements are not purely \textit{true} or \textit{false}, but are associated with confidence values, probability intervals or \textit{degrees of truth} denoting the intrinsic uncertainly of AI problems. It is therefore incorrect to assume that the use of logic is incompatible with uncertainty reasoning or limited to crisp, true or false statements. In its most general form, logic includes fuzzy and many-valued logics and various other forms of non-classical reasoning. We start however with propositional logic. 

Think of propositions as the fundamental building blocks for reasoning. For instance, ``it is raining" is a proposition because its truth can be determined by examining the current weather conditions. We typically use symbols such as $P$, $Q$, or $R$ to represent these propositions.\footnote{Any symbol, including indices, can be used as long as it is clear that they represent a specific proposition.} To combine or modify these propositions, we use logical connectives or operators: AND ($\land$), OR ($\lor$), NOT ($\lnot$), IMPLICATION ($\rightarrow$), and BI-CONDITIONAL ($\leftrightarrow$). For example, if $P$ represents ``it is raining" and $Q$ represents ``I have an umbrella," then $P \land Q$ means ``it is raining AND I have an umbrella". The operators allow us to compose complex relationships among ideas in a precise way.

A syntactically correct expression in logic is said to be a Well-Formed Formula (WFF). A WFF in propositional logic is constructed according to the following rules:
\begin{enumerate}
\item Any atomic proposition (e.g, $P$, $Q$, $R$) is a WFF.
\item If $A$ is a WFF then $\lnot A$ (the negation of $A$) is also a WFF.
\item If $A$ and $B$ are WFFs then $(A \land B)$, $(A \lor B)$, $(A \rightarrow B)$, and $(A \leftrightarrow B)$ are also WFFs.
\item Nothing else is a WFF.
\end{enumerate}

For example, the expression $(P \land Q) \rightarrow R$ is a WFF because it follows these rules: $P$, $Q$, and $R$ are atomic propositions, $(P \land Q)$ is a valid combination using the AND operator, and the entire expression forms a valid implication. On the other hand, expressions like $P \land \lor Q$ are not WFFs because they violate the rules.

Propositional logic is also known as Boolean Logic, named after George Boole, a pioneer in the formalization of logical reasoning. Interestingly, George Boole is the great-great-grandfather of Geoffrey Hinton, a leading figure in the field of neural networks. Boole proposed his \textit{Laws of Thought} using a simplified notation where $1$ and $0$ denote \textit{true} and \textit{false}, respectively. This binary representation aligns naturally with the semantic interpretation of neural networks and fits seamlessly into the reasoning method to be presented in this paper.

By adhering to the rules of WFFs, we ensure that our logical expressions are unambiguous and well-structured (compositional), providing a solid foundation for further exploration of propositional logic and its applications. In the remainder of this paper, unless otherwise specified, we shall use WFF to refer specifically to a subset of WFFs consisting only of formulas constructed using combinations of negation ($\lnot$), conjunction ($\land$), and disjunction ($\lor$). If other logical connectives, such as implication ($\rightarrow$) or bi-conditional ($\leftrightarrow$), are included, we will explicitly clarify this deviation from the specific subset, noting that in Classical Logic $A \leftrightarrow B$ is equavelent to $(A \rightarrow B) \land (B \rightarrow A)$ and that $A \rightarrow B$ is equivalent to $\neg A \lor B$.



\subsection{Illustrating Logical Reasoning with the Sudoku Puzzle}

\begin{figure}[ht]
    \centering
    \includegraphics[width=0.7\textwidth]{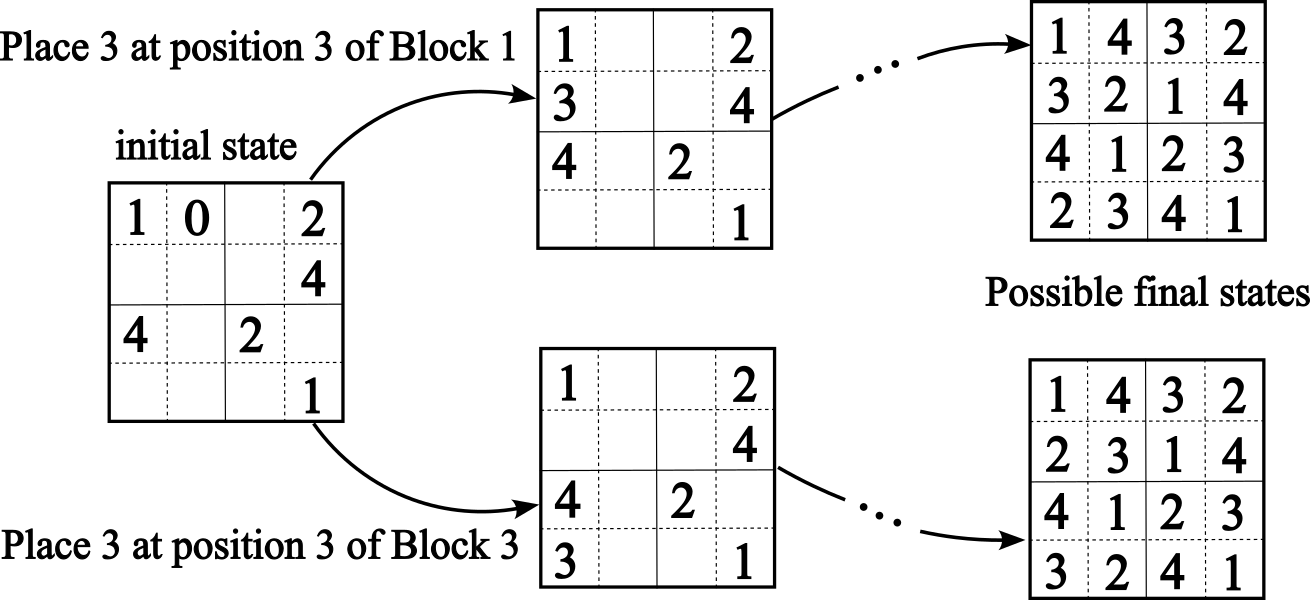}
    \caption{An initial Sudoku board and two branches generated by placing a 3 at position 3 of blocks 1 and 3, respectively, and corresponding final states satisfying the constraints of the game.}
    \label{fig:sudoku_puzzle01}
\end{figure}

Sudoku is more than just a number puzzle (see Figure \ref{fig:sudoku_puzzle01}); it is a gateway to understanding the power of logical thinking. This globally beloved puzzle challenges us to impose order on apparent chaos, using nothing but numbers and logic. At its core, Sudoku is about solving constraints, ensuring that every row, column, and sub-grid (or block) adheres to a simple strict rule (containing one and only one of the elements of a given set). The same principle of constraint satisfaction is a cornerstone of Artificial Intelligence and computational problem-solving. By learning how to express Sudoku’s rules logically, we unlock the secrets of this captivating game and the tools to tackle more complex problem solving. Let’s explore how propositional logic can elegantly capture the rules of Sudoku as a way to illustrate structured reasoning.

For simplicity, we consider a smaller version of Sudoku, using a $4 \times 4$ grid instead of the standard $9 \times 9$. This simplified puzzle divides the board into four $2 \times 2$ blocks or sub-grids, each containing four positions (or cells). Blocks are counted from left to right and top to bottom: block 1 is on top of block 3, and block 2 is on top of block 4. Positions within each block are also counted from left to right and top to bottom. Each cell in the grid must contain a number from 1 to 4, with no repetition allowed in any row, column, or $2 \times 2$ block.\footnote{In the real Sudoku puzzle, each block is $3 \times 3$ and the set of possible elements is \{1,2,...,9\} with the board having 9 blocks in total.} Figure~\ref{fig:sudoku_puzzle01} depicts an example of an initial setting for a Sudoku $4 \times 4$ board, followed by two possible transitions placing number 3 in two possible cells satisfying the constraints. Two possible final states are also shown, each derived from the above two states if every movement satisfies the constraints of the puzzle.

Solving Sudoku involves reasoning about these constraints, making it a good example for introducing logical notation. To model the problem using propositional logic, one can systematically represent the constraints in terms of propositional variables encoding the relationships between numbers, positions, rows, columns and blocks. 
The rules dictate that every row, column and block must include the numbers 1 to 4 exactly once. By encoding the problem in this way, one can use symbolic logical reasoning to systematically explore possible solutions while respecting all constraints. The rules are encoded as follows:







\begin{description}
\item [Logical Variables:] Let the proposition $B_{i,j,k}$ denote that the block $i$ at position $j$ (that is, the cell $(i,j)$) contains the number $k$. Formally, $ B_{i,j,k} $ is true if and only if $ k \in \{1, 2, 3, 4\} $ is in position 
$ j $ of block $ i $, $1 \leq i \leq 4$, $1 \leq j \leq 4$.
        
        
\item [Logical Constraints:] 
The constraints ensure that the numbers are placed correctly according to the rules of Sudoku. These constraints can be grouped into four categories:
\end{description}

\begin{enumerate}
\item Each cell must contain a number (cell $(i,j)$ contains a 1 or a 2 or a 3 or a 4): $B_{i,j,1} \lor B_{i,j,2} \lor B_{i,j,3} \lor B_{i,j,4}$. When needed, we shall write:
 
\[
 \bigvee_{k=1}^4 B_{i,j,k} \quad \text{as shorthand notation for }  B_{i,j,1} \lor B_{i,j,2} \lor B_{i,j,3} \lor B_{i,j,4}.
\]
   
   There cannot be two or more numbers on the same cell\footnote{Notice that $\neg (A \wedge B)$ implies $\neg (A \wedge B \wedge C)$.}: 
   $$
   \neg (B_{i,j,k_1} \wedge B_{i,j,k_2}), \quad \text{for all } k_1 \neq k_2.
   $$
   The above two rules can be written compactly as:
   $$
   \left( \bigvee_{k=1}^4 B_{i,j,k} \right) \wedge \left( \bigwedge_{k_1 < k_2} \neg (B_{i,j,k_1} \wedge B_{i,j,k_2}) \right),
   $$
where $\bigwedge_{i} x_i$ is shorthand for $x_1 \wedge x_2 \wedge ...$ and $k_1 < k_2$ is used to avoid repetition.\footnote{Notice that $A \wedge B$ is logically equivalent to $B \wedge A$.} 

\item Each number appears exactly once per row. For each row across the entire board and each number $ k $, exactly one position in that row must contain $ k $. This is expressed as:
   $$
   \bigvee_{j=1}^4 B_{i,j,k} \quad \text{}
   $$
   
   along with the constraint that there cannot be two or more occurrences of the same number on the same row: 
   $$
   \neg (B_{i,j_1,k} \wedge B_{i,j_2,k}), \quad \text{for all } j_1 \neq j_2.
   $$
   In compact form:
   $$
   \left( \bigvee_{j=1}^4 B_{i,j,k} \right) \wedge \left( \bigwedge_{j_1 < j_2} \neg (B_{i,j_1,k} \wedge B_{i,j_2,k}) \right).
   $$

\item Each number appears exactly once per column. In compact form (as above):

   
   
   $$
   \left( \bigvee_{i=1}^4 B_{i,j,k} \right) \wedge \left( \bigwedge_{i_1 < i_2} \neg (B_{i_1,j,k} \wedge B_{i_2,j,k}) \right).
   $$

\item Each number appears exactly once per block. For each $ 2 \times 2 $ block and each number $ k $, exactly one position within the block must contain $ k $. For example, for the top-left block:
   $$
   \bigvee_{(i,j) \in \{(1,1), (1,2), (2,1), (2,2)\}} B_{i,j,k},
   $$
   
   along with the constraint:
   $$
   \neg (B_{i_1,j_1,k} \wedge B_{i_2,j_2,k}), \text{for all distinct pairs } (i_1,j_1) \neq (i_2,j_2).
   $$
   In compact form:
   $$
   \left( \bigvee_{(i,j) \in \text{block}} B_{i,j,k} \right) \wedge \left( \bigwedge_{(i_1,j_1) < (i_2,j_2)} \neg (B_{i_1,j_1,k} \wedge B_{i_2,j_2,k}) \right).
   $$
\end{enumerate}


The complete set of constraints for the $4 \times 4$ Sudoku puzzle is the conjunction of all the above conditions over all cells, rows, columns and blocks. This logical formula guarantees that every number appears exactly once in each row, column, and block, satisfying the rules of Sudoku. It also provides a systematic framework for reasoning about the puzzle.

\begin{example}
For block 1, position 1, we have:

\begin{itemize}
    \item [] $B_{1,1,1} \lor B_{1,1,2} \lor B_{1,1,3} \lor B_{1,1,4}$
    \item [] $\neg B_{1,1,1} \lor \neg B_{1,1,2}$
    \item [] $\neg B_{1,1,1} \lor \neg B_{1,1,3}$
    \item [] $\neg B_{1,1,1} \lor \neg B_{1,1,4}$
    \item [] $\neg B_{1,1,2} \lor \neg B_{1,1,3}$
    \item [] $\neg B_{1,1,2} \lor \neg B_{1,1,4}$
    \item [] $\neg B_{1,1,3} \lor \neg B_{1,1,4}$
\end{itemize}
\end{example}

Some observations about this representation:
\begin{itemize}
    \item This notation provides a framework whereby each possible combination of $B$ with indices is assigned to \textit{True} or \textit{False}.
    \item Each rule above is called a \textit{clause} (a disjunction of logic literals) and the complete set of clauses would be significantly larger to cover all rows, columns and blocks.
    \item This representation can be used as input to a satisfiability (SAT) solver to find solutions to the Sudoku puzzle, that is, assignments of truth-values \textit{True} or \textit{False} to each literal that will provably satisfy the puzzle's constraints.
\end{itemize}

This Boolean logic representation allows us to express the Sudoku problem as a set of constraints that must be satisfied simultaneously. By finding a truth assignment to the variables that satisfy all the clauses, we determine a valid solution to the Sudoku puzzle.

\subsection{Sudoku with Strategies of Sampling}


\begin{enumerate} 
\item  Reasoning Strategy based on Unused Numbers:

To control which number to pick based on the bank of numbers not yet placed on the board, let us illustrate how additional constraints may be introduced that ensure unused numbers are considered first. A strategy such as this could be learned from observation of game plays as well as specified by hand.

For each empty cell \( (i, j) \), define \( U(i, j) \) as the set of numbers \( k \) such that \( k \) is not already used in the corresponding row, column or block of cell \( (i, j) \).

The constraint ensuring the selection of an unused number \( k \) can be expressed as:
\[
\bigvee_{k \in U(i, j)} B_{i,j,k}
\]

where \( U(i, j) \) is defined as:
\[
U(i, j) = \{ k \mid k \notin \{ B_{i,j',k'} \mid j' \neq j \} \land k \notin \{ B_{i',j,k'} \mid i' \neq i \} 
\]
\[
\land k \notin \{ B_{i',j',k'} \mid (i', j') \in \text{block}(i, j) \} \}.
\]

Here, \( \text{block}(i, j) \) denotes the set of positions in the same block as \( (i, j) \).

\item Priority Constraint for Unused Numbers:

To prioritize the use of unused numbers, we can add a preference rule that assigns higher priority to considering numbers from \( U(i, j) \) ahead of other possibilities.

Formally, let \( P(i, j, k) \) represent the priority of placing number \( k \) in cell \( (i, j) \). The priority can be defined as:
\[
P(i, j, k) = 
\begin{cases} 
1 & \text{if } k \in U(i, j) \\
0 & \text{otherwise}
\end{cases}
\]

The constraint ensuring that the highest priority is given to unused numbers can be expressed as:
\[
\bigvee_{k \in U(i, j)} (P(i, j, k) \wedge B_{i,j,k})
\]

\end{enumerate}


The complete set of logical constraints for the 4x4 Sudoku puzzle now includes the original Sudoku constraints along with additional reasoning strategies that prioritize the use of unused numbers. These constraints ensure that every number appears exactly once in each row, column, and block while also guiding the generation of solutions (that is, the assignment of truth-values to the literals) by leveraging the bank of unused numbers. By incorporating these, the Sudoku solving process becomes systematic and more efficient as it should reduce the likelihood of the process getting stuck and having to backtrack when searching for a solution, or analogously in the case of a neural network getting stuck in local minima.



\subsection{Restricted Boltzmann Machines}

\label{dnf_rbm}
An RBM \cite{Smolensky_1995} is a two-layer neural network with bidirectional (symmetric) connections,
which is characterised by a function called the energy of the RBM:

{
\begin{equation}
\label{eq:rbm_en}
  \En(\vt{x},\vt{h}) = -\sum_{i,j} w_{ij}x_ih_j - 
\sum_{i}a_ix_i - \sum_jb_jh_j
\end{equation}
}

\noindent where $a_i$ and $b_j$ are the biases of input unit $x_i$ and hidden unit $h_j$, respectively, and $w_{ij}$ is the connection weight between $x_i$ and $h_j$. This RBM represents a joint probability distribution $p(\vt{x},\vt{h})= \frac{1}{Z}e^{-\frac{1}{\tau}\En(\vt{x},\vt{h})}$ where $Z=\sum_{\vt{x}\vt{h}}e^{-\frac{1}{\tau}\En(\vt{x},\vt{h})}$ is the partition function and parameter $\tau$ is called the temperature of the RBM, $\vt{x} = \{x_i\}$ is the set of visible units and $\vt{h} = \{h_j\}$ is the set of hidden units of the RBM.

Training RBMs normally makes use of the Contrastive Divergence learning algorithm \cite{Hinton_2006}, whereby each input vector from the training set is propagated to the hidden layer of the network and back to the input a number of times ($n$) using a probabilistic selection rule to decide at each time whether or not a neuron should be activated (with activation value in $\{0,1\}$). The weight assigned to the connection between input neuron $x_i$ and hidden neuron $h_j$ is adjusted according to a simple update rule based on the difference between the value of $x_i h_j$ at time $1$ and time $n$. More precisely, $\Delta W_{ij}= \eta ((x_i h_j)_1 - (x_i h_j)_n)$, where $\eta$ is a learning rate (a small positive real number).

%% file: AAAI.tex
The content of this section is based on \cite{Tran_Garcez_2023}.

Over the years, many neurosymbolic approaches have used a form of knowledge representation based on \emph{if-then} rules \cite{Towel_1994,Franca_2014,Son_2018,Evans_18,Yang_2017,Robin_2018,Tran_2021}, written $B \lif A$ (make $B$ $True$ if $A$ is $True$) to distinguish from classical implication ($A \rightarrow B$). Under the convention that $1$ represents $True$ and $0$ represents $False$, given $B \lif A$ and input $1$ to neuron $A$, a neurosymbolic system would infer that neuron $B$ should have activation value approximately $1$. Given input $0$ to neuron $A$, it would infer that $B$ should have activation approximately $0$. 




Logical Boltzmann Machines (LBM) allow for a richer representation than \emph{if-then} rules by using full propositional logic. Next, we review LBM's immediate related work, define a mapping from any logical formulae to LBMs, and describe how reasoning takes place by sampling and energy minimization. We also evaluate scalability of reasoning in LBM and learning by combining knowledge and data, evaluating results on benchmarks in comparison with a symbolic, another neurosymbolic and a neural network-based approach. 

\subsection{Related Work}
\label{sec:background}
One of the earliest work on the integration of neural networks and symbolic knowledge is known as KBANN (Knowledge-based Artificial Neural Network \cite{Towel_1994}), which encodes {\it if-then} rules into a hierarchical multilayer perceptron. In another early approach \cite{Garcez_2001}, a single-hidden layer recurrent neural network is proposed to support logic programming rules. An extension of that approach to work with first-order logic programs, called Connectionist Inductive Logic Programming (CILP++) \cite{Franca_2014}, uses the concept of {\it propositionalisation} from Inductive Logic Programming (ILP), whereby first-order variables can be treated as propositional atoms in the neural network. Also based on first-order logic programs, \cite{Evans_18} propose a differentiable ILP approach that can be implemented by neural networks, and \cite{Cohen_2017} maps stochastic logic programs into a differentiable function also trainable by neural networks. These are all supervised learning approaches.

Early work in neurosymbolic AI has also shown a correspondence between propositional logic and symmetrical neural networks \cite{Pinkas_1991b}, in particular Hopfield networks, which nevertheless did not scale well with the number of variables. Among unsupervised learning approaches, Penalty Logic \cite{Pinkas_1995} was the first work to integrate nonmonotonic logic in the form of weighted if-then rules into symmetrical neural networks. However, Penalty Logic required the use of higher-order Hopfield networks, which can be difficult to construct\footnote{Building such higher-order networks requires transforming the energy function into quadratic form by adding hidden variables not present in the original logic formulae.} and inefficient to train with the learning algorithm for Boltzmann machines. More recently, several attempts have been made to extract and encode symbolic knowledge into RBMs trained with the more efficient Contrastive Divergence learning algorithm \cite{Leo_2011, Son_2018}. Such approaches explored the structural similarity between symmetric networks and logical rules with bi-conditional implication but do not have a proof of soundness. By contrast, and similarly to Penalty Logic, LBM is provably equivalent to the logic formulae encoded in the RBM. Differently from Penalty Logic, LBM does not require the use of higher-order networks.

Alongside the above approaches, which translate symbolic representations into neural networks (normally if-then rules translated into a feedforward or recurrent network), there are hybrid approaches that combine neural networks and symbolic AI systems as communicating modules of a neurosymbolic system. These include DeepProbLog \cite{Robin_2018} and Logic Tensor Networks (LTN) \cite{Serafini_2016}. DeepProbLog adds a neural network module to probabilistic logic programming such that an atom of the logic program can be represented by a network module. LTN and various approaches derived from it use real-valued logic to constrain the loss function of the neural network given statements in firt-order logic. Both DeepProbLog and LTNs use backpropagation, differently from the approach adopted here which uses Contrastive Divergence.  






Finally, approaches focused on reasoning include SAT solving using neural networks. In \cite{SAT1, SAT2}, the maximum satisfiability problem is mapped onto Boltzmann machines and higher-order Boltzmann machines, which are used to solve the combinatorial optimization task in parallel, similarly to \cite{Pinkas_1991b}. In \cite{satnet}, the SAT problem is redefined as a soft (differentiable) task and solved approximately by deep networks with the objective of integrating logical reasoning and learning, as in the case of the approaches discussed earlier. This soft version of the SAT problem is therefore different from the satisfiability problem. A preliminary evaluation of our approach in comparison with symbolic SAT solvers shows that our approach allows the use of up to approximately 100 variables. This is well below the capability of symbolic SAT solvers. A way of improving the performance of neural SAT solvers may well be to consider approximate solutions as done by soft SAT solvers, including neuroSAT \cite{selsam2018learning}. Although still not beating SAT solvers, neuroSAT showed promise at addressing out-of-distribution learning after training on random SAT problems. 

In our experiments on learning, the focus is on benchmark neurosymbolic AI tasks with available data and knowledge, obtained from \cite{Franca_2014}. We therefore compare LBM with a state-of-the-art ILP symbolic system ALEPH \cite{aleph}, standard RBMs as a purely-neural approach closest to LBM, and with CILP++ as a neurosymbolic system. It is worth noting, however, that CILP++ is a neurosymbolic system for supervised learning while LBMs use unsupervised learning, and it is worth investigating approaches for semi-supervised learning and other combinations of such systems. Further comparisons and evaluations on both reasoning and learning are underway. 

\vspace{-0.3cm}
\subsection{Knowledge Representation in RBMs}
\label{theory}

Before we present LBM, let's contrast the simple $B \lif A$ example used earlier with classical logic. Given $A \rightarrow B$ as knowledge\footnote{In classical logic, $A \rightarrow B$ is equivalent to $\neg A \vee B$, i.e. \textit{True} if $A$ is \textit{False} regardless of the truth-value of $B$.}, if neuron $A$ is assigned input value $1$ in the corresponding neurosymbolic network, we expect the network to converge to a stable state where neuron $B$ has value approximately $1$, similarly to the example seen earlier. This is because the truth-value of WFF $A \rightarrow B$ is \textit{True} given an assignment of truth-values \textit{True} to its constituent literals $A$ and $B$. Now, $A \rightarrow B$ is \textit{False} when $A$ is \textit{True} and $B$ is \textit{False}. If neuron $B$ is assigned input $0$, we expect the network to converge to a stable state where $A$ is approximately $0$ ($A \rightarrow B$ is \textit{True} when $A$ is \textit{False} and $B$ is \textit{False}). What if $A$ is assigned input $0$ (or $B$ is assigned input $1$)? In these cases, $A \rightarrow B$ is satisfied if $B$ is either $1$ or $0$ (or if $A$ is either $1$ or $0$). Differently from $B \lif A$, the network will converge to one of the two options that satisfy the formulae. 

From this point forward, unless stated otherwise, we will treat assignments of truth-values to logical literals and binary input vectors denoting the activation states of neurons indistinguishably. 


\begin{definition}
\label{equival}
Let $s_\varphi(\vt{x}) \in \{0,1\}$ denote the truth-value of a WFF $\varphi$ given an assignment of truth-values $\vt{x}$ to the literals of $\varphi$, where truth-value $True$ is mapped to 1 and truth-value $False$ is mapped to 0. Let $\En(\vt{x},\vt{h})$ denote the energy function of an energy-based neural network $\mathcal{N}$ with visible units $\vt{x}$ and hidden units $\vt{h}$. $\varphi$ is said to be \emph{equivalent} to $\mathcal{N}$ if and only if for any assignment of values to $\vt{x}$ there exists a function $\psi$ such that $s_\varphi(\vt{x}) = \psi(\En(\vt{x},\vt{h}))$.


\end{definition} 
Definition \ref{equival} is similar to that of Penalty Logic \cite{Pinkas_1995}, where all assignments of truth-values satisfying a WFF $\varphi$ are mapped to global minima of the energy function of network $\mathcal{N}$. In our case, by construction, assignments that do not satisfy the WFF will, in addition, be mapped to maxima of the energy function. To see how this is the case, it will be useful to define strict and full DNFs, as follows.

\begin{definition}
A \emph{strict DNF} (SDNF) is a DNF with at most one conjunctive clause (a conjunction of literals) that maps to $True$ for any choice of assignment of truth-values $\vt{x}$. 
A \emph{full DNF} is
    a DNF where each propositional variable (a positive or negative literal) must appear at least once in every
    conjunctive clause (sometimes called a canonical DNF). 
\end{definition}

For example, to turn DNF $A \vee B$ into an equivalent full DNF, one needs to map it to $(A \wedge \neg B) \vee (\neg A \wedge B) \vee (A \wedge B)$, according to the truth-table for $A \vee B$. For any given assignment of truth-values to $A$ and $B$, at most one of the above three conjunctive clauses will be $True$, by definition of the truth-table. Not every SDNF is also a full DNF though, e.g. $(a \wedge b) \vee \neg b$ is a SDNF that is not a full DNF. 

\begin{lemma}
\label{lem:wff2ecs}
Let $\mathcal{S}_{T_j}$ denote the set of indices of the positive literals $\pr{x}_t$ in a conjunctive clause $j$. Let $\mathcal{S}_{K_j}$ denote the set of indices of the negative literals $\pr{x}_k$ in $j$. Any SDNF $\varphi \equiv \bOr_j (\bAnd_{t}
\pr{x}_t \fzand \bAnd_{k} \neg \pr{x}_{k})$ can
be mapped onto an energy function:
 
\begin{equation*}
\En(\vt{x}) = -\sum_{j} (\prod_{t \in \mathcal{S}_{T_j}} x_t \prod_{k \in \mathcal{S}_{K_j}} (1 - x_{k})).
\end{equation*}

\end{lemma}
\textbf{Proof:}
Each conjunctive clause $\bAnd_{t}
\pr{x}_t \fzand \bAnd_{k} \neg \pr{x}_{k}$ in $\varphi$
corresponds to the product $\prod_{t} x_t \prod_{k} (1-x_{k})$ which maps to $1$ if and only if $x_t$ is $True$ ($x_t=1$) and $x_{k}$ is $False$ ($x_k=0$) for all $t \in \mathcal{S}_{T_j}$ and $k \in \mathcal{S}_{K_j}$. Since
$\varphi$ is SDNF, $\varphi$ is $True$ if and only if one conjunctive
clause is $True$ and $\sum_{j}( \prod_{t\in
  \mathcal{S}_{T_j}} x_t \prod_{k \in \mathcal{S}_{K_j}} (1-x_{k}))=1$. Hence, the neural network with energy
function $\En$ is such that $s_\varphi(\vt{x}) = -
\En(\vt{x})$.
\qed

\begin{theorem}
\label{theorem:prop_rbm} 
 Any SDNF $\varphi \equiv \bOr_j (\bAnd_{t}
 \pr{x}_t \fzand \bAnd_{k} \neg \pr{x}_{k})$ can
 be mapped onto an RBM with energy function:
\begin{equation}
\En(\vt{x},\vt{h}) =
 -\sum_jh_j(\sum_{t \in \mathcal{S}_{T_j}} x_t - \sum_{k \in
   \mathcal{S}_{K_j}}x_{k} - |\mathcal{S}_{T_{j}}| + \epsilon),
\end{equation}
\noindent such that $s_\varphi(\vt{x}) = - \En(\vt{x})$, where $0<\epsilon<1$ and $|\mathcal{S}_{T_{j}}|$ is the number of positive literals in conjunctive clause $j$ of $\varphi$.


\end{theorem}

\textbf{Proof:} 
Lemma \ref{lem:wff2ecs} states that any SDNF $\varphi$ can be
mapped onto energy function $\En = -\sum_{j} (\prod_{t\in \mathcal{S}_{T_j}}
x_t \prod_{k \in \mathcal{S}_{K_j}} (1-x_{k})) $. 
For each expression
$\tilde{e}_j(\vt{x}) = -\prod_{t \in \mathcal{S}_{T_j}} x_t \prod_{k \in \mathcal{S}_{K_j}}
(1-x_{k})$, we define an energy expression associated with hidden unit
$h_j$ as $e_j(\vt{x},h_j) = -h_j(\sum_{t \in \mathcal{S}_{T_j}} x_t
- \sum_{k \in \mathcal{S}_{K_j}}x_{k} - |\mathcal{S}_{T_{j}}| + \epsilon)$. The term $e_j(\vt{x},h_j)$ is minimized with value $-\epsilon$ when $h_j=1$, written $min_{h_j}(e_j(\vt{x},h_j)) = -\epsilon$. This is because $-(\sum_{t \in \mathcal{S}_{T_j}}
x_t - \sum_{k \in \mathcal{S}_{K_j}} x_{k} -|\mathcal{S}_{T_{j}}| + \epsilon) = -\epsilon$ if and only if $x_t=1$ and
$x_{k}=0$ for all $t \in \mathcal{S}_{T_j}$ and $k \in \mathcal{S}_{K_j}$.
Otherwise, $-(\sum_t x_{t \in \mathcal{S}_{T_j}}
- \sum_{k \in \mathcal{S}_{K_j}} x_{k} -|\mathcal{S}_{T_{j}}| +\epsilon) >0$ and  $min_{h_j}(e_j(\vt{x},h_j)) =
0$ with $h_j=0$. By repeating this process for each
$\tilde{e}_j(\vt{x})$ we obtain that the energy function
{
 $
   \label{eq:prop_rbm_en}
   \En(\vt{x},\vt{h}) = -\sum_jh_j(\sum_{t \in \mathcal{S}_{T_j}} x_t - \sum_{k \in \mathcal{S}_{K_j}}x_{k}  -  |\mathcal{S}_{T_{j}}| + \epsilon)  
 $
 }
is such that $s_\varphi(\vt{x}) = -\frac{1}{\epsilon} min_{\vt{h}}\En(\vt{x},\vt{h})$.
\qed



\label{RepCap}
It is well-known that any WFF $\varphi$ can be converted into DNF. Then, if $\varphi$ is not SDNF, by definition there is more than one conjunctive clause in $\varphi$ that map to $True$ when $\varphi$ is satisfied. This group of conjunctive clauses can always be converted into a full DNF according to its truth-table. By definition, any such full DNF is also a SDNF. Therefore, any WFF can be converted into SDNF. From Theorem \ref{theorem:prop_rbm}, it follows that any WFF can be represented by the energy function of an RBM. The conversion of WFFs into full DNF can be computationally expensive. Sometimes, the logic is provided already in canonical DNF form or in Conjunctive Normal Form (CNF), i.e. conjunctions of disjunctions. We will see later that any WFF expressed in CNF can be converted into an RBM's energy function efficiently without the need to convert into SDNF first. This covers the most common forms of propositional knowledge representation. Next, we describe a method for converting logical formulae into SDNF, which we use in the empirical evaluations that will follow. Consider a clause $\gamma$ such that:

{
\begin{equation}
\label{eq:cl}
\gamma \equiv \bOr_{t\in \mathcal{S}_T} \neg \pr{x}_t \fzor \bOr_{k\in \mathcal{S}_K} \pr{x}_k
\end{equation}
}

\noindent where $\mathcal{S}_T$ now denotes the set of indices of the negative literals, and $\mathcal{S}_K$ denotes the set of indices of the positive literals in the clause (dually to the conjunctive clause case). Clause $\gamma$ can be rearranged into $\gamma \equiv \gamma' \fzor \pr{x}'$, where $\gamma'$ is obtained by removing $\pr{x}'$ from $\gamma$ ($\pr{x}'$ can be either $\neg \pr{x}_t$ or $\pr{x}_k$ for any $t \in \mathcal{S}_T$ and $k \in \mathcal{S}_K$). We have: 
\begin{equation}
\label{conj_ext}
\gamma  \equiv (\neg \gamma' \fzand \pr{x}') \fzor \gamma'
\end{equation}
because $(\neg \gamma' \fzand \pr{x}') \fzor \gamma' \equiv (\gamma' \fzor \neg \gamma') \fzand (\gamma' \fzor \pr{x}') \equiv True \fzand (\gamma' \fzor \pr{x}')$. By De Morgan's law ($\neg( \pr{a}\fzor \pr{b})\equiv \neg \pr{a}\fzand\neg\pr{b}$), we can always convert $\neg \gamma'$ (and therefore $\neg \gamma' \fzand \pr{x}'$) into a conjunctive clause. 

By applying \eqref{conj_ext} repeatedly, each time we eliminate a variable out of the clause by moving it into a new conjunctive clause. Given an assignment of truth-values, either the clause $\gamma'$ will be \textit{True} or the conjunctive clause ($\neg \gamma' \fzand \pr{x}'$) will be \textit{True}, e.g. $a \vee b \equiv a \vee (\neg a \wedge b)$. Therefore, the SDNF for clause $\gamma$ in Eq. \eqref{eq:cl} is:
{
\begin{equation}
\label{short_horn_dnf}
\bOr_{p \in \mathcal{S}_T \cup \mathcal{S}_K} (\bAnd_{t\in\mathcal{S}_T\backslash p} \pr{x}_t \fzand  \bAnd_{k\in\mathcal{S}_K\backslash p} \neg\pr{x}_k \fzand \pr{x}'_{p})
\end{equation}
 }

\noindent where $\mathcal{S} \backslash p$ denotes a set $\mathcal{S}$ from which element $p$ has been removed. If $p\in\mathcal{S}_T$ then $\pr{x}'_p\equiv\neg\pr{x}_p$. Otherwise, $\pr{x}'_p\equiv\pr{x}_p$. 
As an example of the translation into SDNF, consider the translation of an \textit{if-then} statement (logical implication) below.

\begin{example} Translation of if-then rules into SDNF. Consider the formula $\gamma \equiv (x_1 \wedge x_2 \wedge \neg x_3) \rightarrow y$. Using our notation:

{
\begin{equation}
\label{horn}
\gamma \equiv (\bAnd_{t\in \{1,2\}} \pr{x}_{t} \fzand \bAnd_{k \in \{3\}} \neg \pr{x}_k) \rightarrow \pr{y} 
\end{equation}
}

Converting to DNF:

{
\begin{equation}
\label{logimp_dnf}
(\pr{y} \fzand \bAnd_{t\in \{1,2\}} \pr{x}_t \fzand \bAnd_{k\in \{3\}}  \neg \pr{x}_k) \fzor \bOr_{t\in \{1,2\}} \neg \pr{x}_t \fzor \bOr_{k\in \{3\}} \pr{x}_k
\end{equation}
}

Applying the variable elimination method to the clause $\neg \pr{x}_1 \fzor \neg \pr{x}_2 \fzor \pr{x}_3$, we obtain the SDNF for $\gamma$:

{
\begin{equation}
\label{short_horn_dnf2}
\begin{aligned}
(\pr{y} \fzand \bAnd_{t\in \mathcal{S}_T} \pr{x}_t  \bAnd_{k\in \mathcal{S}_K} \neg \pr{x}_k) \fzor 
(\neg \pr{x}_1) \fzor (\pr{x}_1 \wedge \neg \pr{x}_2) \fzor ( \pr{x}_1 \wedge \pr{x}_2 \wedge \pr{x}_3)
\end{aligned}
\end{equation}
 }
\end{example}

\subsection{Reasoning in RBMs}
\label{sub:reasoning}
We have seen how propositional logic formula can be mapped onto the energy functions of RBMs. In this section, we discuss the deployment of such RBMs for logical reasoning.

\subsubsection{Reasoning as Sampling} There is a direct relationship between inference in RBMs and logical satisfiability, as follows.

\begin{lemma}
  \label{prop:gibbs_sat}
Let $\mathcal{N}$ be an RBM with energy function $E$. Let $\varphi$ be a WFF such that $s_\varphi(\vt{x}) = - \En(\vt{x})$. Let $\mathcal{A}$ be a set of indices of variables in $\varphi$ that have been assigned to either \textit{True} or \textit{False}. We use $\vt{x}_\mathcal{A}$ to denote the set $\{x_\alpha|\alpha \in \mathcal{A}\}$). Let $\mathcal{B}$ be a set of indices of variables that have not been assigned a truth-value in $\varphi$. We use $\vt{x}_\mathcal{B}$ to denote $ \{x_\beta|\beta \in \mathcal{B}\}$). Performing Gibbs sampling on $\mathcal{N}$ given $\vt{x}_\mathcal{A}$ is equivalent to searching for an assignment of truth-values for $\vt{x}_\mathcal{B}$ that satisfies $\varphi$.
\end{lemma}

\textbf{Proof:}
 Theorem \ref{theorem:prop_rbm} has shown that the assignments of truth-values to $\varphi$ are partially ordered according to the RBM's energy function such that the models of $\varphi$ (mapping $\varphi$ to 1) correspond to minima of the energy function. We say that the satisfiability of $\varphi$ is inversely proportional to the RBM's rank function.\footnote{When the satisfiability of $\varphi$ is maximum ($s_\varphi(\vt{x}) = 1$) ranking the output of $-\En(\vt{x})$ produces the highest rank.} A value of $\vt{x}_\mathcal{B}$ that minimises the energy function also maximises satisfiability: $
   s_\varphi(\vt{x}_\mathcal{B},\vt{x}_\mathcal{A}) \propto -
   min_{\vt{h}}\En(\vt{x}_{\mathcal{B}},\vt{x}_{\mathcal{A}},\vt{h})$ because:
 \begin{equation}
 \begin{aligned}
\vt{x}_\mathcal{B}^{*} = \argmin_{\vt{x}_{\mathcal{B}_\vt{h}}} \En(\vt{x}_\mathcal{B}, \vt{x}_\mathcal{A}, \vt{h})
 = \argmax_{\vt{x}_\mathcal{B}} (s_\varphi(\vt{x}_\mathcal{B},\vt{x}_\mathcal{A}))
 \end{aligned}
 \end{equation}
 
We can consider an iterative process to search for truth-values $\vt{x}_\mathcal{B}^*$ by minimising an RBM's energy function. This can be done using gradient descent or contrastive divergence with Gibbs sampling. The goal is to update the values of $\vt{h}$ and then $\vt{x}_\mathcal{B}$ in parallel until convergence to minimise $\En(\vt{x}_\mathcal{B},\vt{x}_\mathcal{A},\vt{h})$ while keeping the other variables ($\vt{x}_\mathcal{A}$) fixed. The gradients amount to:

 \begin{equation}
   \label{eq:en_grad}
   \begin{aligned}
     \frac{\partial -\En(\vt{x}_\mathcal{B},\vt{x}_\mathcal{A},\vt{h})}{\partial h_j} &= \sum_{i\in \mathcal{A}\cup\mathcal{B}} x_iw_{ij} + \theta_j\\
     \frac{\partial -\En(\vt{x}_\mathcal{B},\vt{x}_\mathcal{A},\vt{h})}{\partial x_\beta} &= \sum_j h_jw_{\beta j} + theta_\beta
    \end{aligned}
 \end{equation}
 
 In the case of Gibbs sampling, given the assigned variables $\vt{x}_\mathcal{A}$, the process starts with a random initialization of $\vt{x}_\mathcal{B}$ and proceeds to infer values for the hidden units $h_j$ and then the unassigned variables $x_\beta$ in the visible layer of the RBM, using the conditional distributions $h_j \sim p(h_j|\vt{x})$ and $x_\beta \sim p(x_\beta|\vt{h})$, respectively, where $\vt{x}=\{\vt{x}_\mathcal{A},\vt{x}_\mathcal{B}\}$ and:
 
 \begin{equation}
   \label{eq:gibbs}
   \centering
   \begin{aligned}
     p(h_j|\vt{x}) & = \frac{1}{1+e^{-\frac{1}{\tau}\sum_i x_i w_{ij}+\theta_j}} = \frac{1}{1+e^{-\frac{1}{\tau} \frac{\partial -\En(\vt{x}_\mathcal{B},\vt{x}_\mathcal{A},\vt{h})}{\partial h_j}}}\\ \\
     p(x_\beta|\vt{h}) & = \frac{1}{1+e^{-\frac{1}{\tau}\sum_j h_j w_{\beta j}+\theta_\beta}} = \frac{1}{1+e^{-\frac{1}{\tau} \frac{\partial -\En(\vt{x}_\mathcal{B},\vt{x}_\mathcal{A},\vt{h})}{\partial x_\beta}}}\\
   \end{aligned}
  \end{equation}
 
It can be seen from Eq.\eqref{eq:gibbs} that the distributions are monotonic functions of the negative energy's gradient over $\vt{h}$ and $\vt{x}_\mathcal{B}$. Therefore, performing Gibbs sampling on them can be seen as moving towards a local minimum that is equivalent to an assignment of truth-values that satisfies $\varphi$. Each step of Gibbs sampling, calculating $\vt{h}$ and then $\vt{x}$ to reduce the energy, should intuitively generate an assignment of truth-values that gets closer to satisfying the formula $\varphi$.
\qed

\subsubsection{Reasoning as Lowering Free Energy} \label{cond_dis} 
\begin{figure}[ht]
\centering
\includegraphics[width=0.55\textwidth]{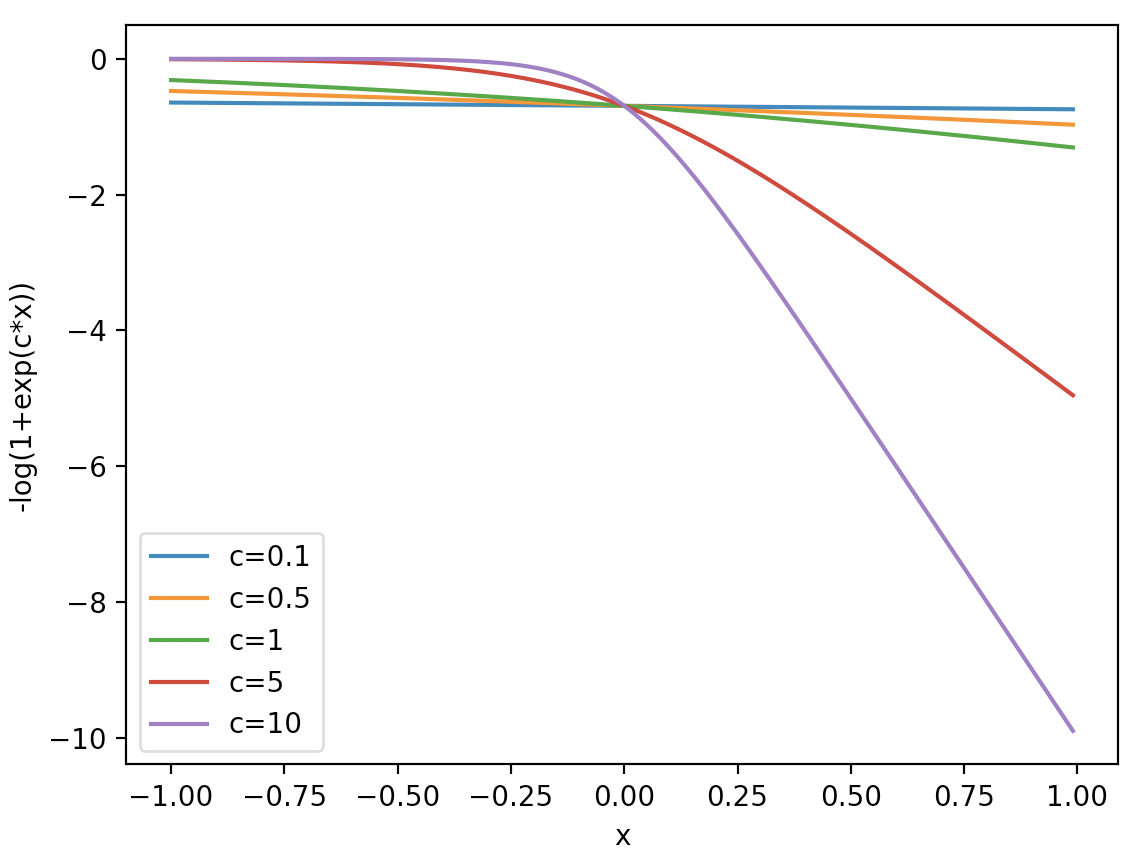}
\caption{Free energy term $-\log(1+e^{cx})$ for different confidence values $c$.}
\label{fig:softplus}
\end{figure}

When the number of unassigned variables is not large, it should be possible to calculate the above probabilities directly. In this case, one can infer the assignments of $\vt{x}_\mathcal{B}$ using the conditional distribution:
\begin{equation}
    P(\vt{x}_\mathcal{B}|\vt{x}_\mathcal{A}) = \frac{e^{-\mathcal{F}_{\mathcal{B}}(\vt{x}_\mathcal{A},\vt{x}_\mathcal{B})}}{\sum_{\vt{x}'_\mathcal{B}}e^{\mathcal{F}_{\mathcal{B}}}(\vt{x}_\mathcal{A},\vt{x}'_\mathcal{B})}
\end{equation}

\noindent where $\mathcal{F}_\mathcal{B}= - \sum_j (- \log(1+e^{(c \sum_{i\in \mathcal{A} \cup \mathcal{B}} w_{ij}x_i + \theta_j)}))$ is known as the free energy; $\vt{x}'_\mathcal{B}$ denotes all the combinations of truth-value assignments to the literals in $\vt{x}_\mathcal{B}$, and $c$ is a non-negative real number that we call a {\it confidence value}. The free energy term $- \log(1+e^{(c \sum_{i\in \mathcal{A} \cup \mathcal{B}} w_{ij}x_i + \theta_j)})$ is a negative softplus function scaled by $c$ as shown in Figure \ref{fig:softplus}. It returns a negative output for a positive input and a close-to-zero output for a negative input. 

 Each free energy term is associated with a conjunctive clause in the SDNF through the weighted sum $\sum_{i\in \mathcal{A} \cup \mathcal{B}} w_{ij}x_i + \theta_j$. Therefore, if a truth-value assignment of $\vt{x}_\mathcal{B}$ does not satisfy the formula $\varphi$, all energy terms will be close to zero. When $\varphi$ is satisfied, one free energy term will be $-\log(1+e^{c\epsilon})$, for a choice of $0<\epsilon<1$ from Theorem \ref{theorem:prop_rbm}. Thus, the more likely that a truth assignment is to satisfying the formula, the lower the free energy. Formally:
 
 \begin{equation}
\label{sat_erank}
s_\varphi(\vt{x}) = -\frac{1}{c\epsilon}\text{min}_{\vt{h}} E(\vt{x},\vt{h}) = \lim_{c\rightarrow \infty} -\frac{1}{c\epsilon} \mathcal{F}(\vt{x})
\end{equation}

\begin{figure}[ht]
    \centering
    \begin{subfigure}{0.4\textwidth}
    \includegraphics[width=1\textwidth]{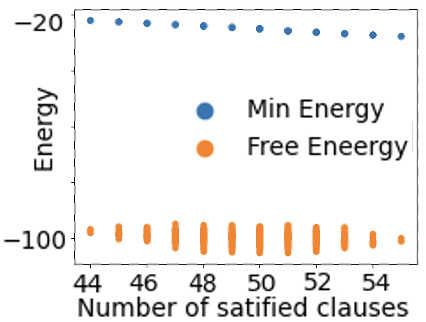}
    \caption{(c=1)}
    \label{enc1}
    \end{subfigure}
    \begin{subfigure}{0.4\textwidth}
    \includegraphics[width=1\textwidth]{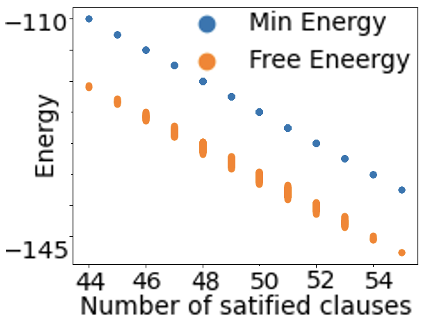}
    \caption{(c=5)}
    \label{enc5}
    \end{subfigure}
    \begin{subfigure}{0.4\textwidth}
    \includegraphics[width=1\textwidth]{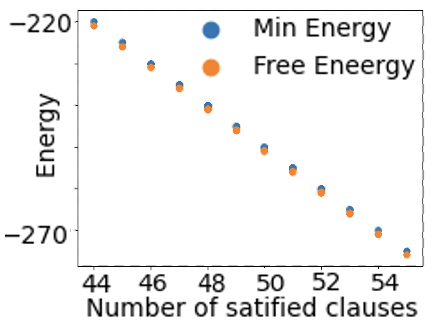}
    \caption{(c=10)}
    \label{enc10}
    \end{subfigure}
    \caption{Linear correlation between satisfiability of a CNF and minimization of the free energy function for various confidence values $c$. Source: \cite{Tran_Garcez_2023}.}
    \label{fig:dnf_exp}
\end{figure}

Figure \ref{fig:dnf_exp} shows the average values of the energy function and free energy for CNFs with 55 clauses as the number of satisfied clauses increases. The CNF is satisfied if and only if all 55 clauses are satisfied. As can be seen, the relationships are linear. Minimum energy and free energy values converge with an increasing value of $c$.

 \subsection{Logical Boltzmann Machines}
 \label{LBMsection}
 
We are now in position to present a translation algorithm to build an RBM from logical formulae. The energy function of the RBM will be derived based on Theorem \ref{theorem:prop_rbm} given a formula in SDNF. The weights and biases of the RBM will be obtained from the energy function $E(\pr{x},\pr{h})=-(\sum_i \theta_i \pr{x}_i + \sum_j \theta_j\pr{h}_j + \sum_{ij} \pr{x}_iW_{ij}\pr{h}_j
 )$, where $\theta_i$ are the biases of the visible units, $\theta_j$ are the biases of the hidden units, and $W_{ij}$ is the symmetric weight between a visible and a hidden unit. For each conjunctive clause in the formula of the form $\bAnd_{t\in\mathcal{S}_T} \pr{x}_t \fzand  \bAnd_{k\in\mathcal{S}_K}\ \neg\pr{x}_k$, we create an energy term $-h_j(\sum_{t\in\mathcal{S}_T}x_t - \sum_{k\in\mathcal{\
S}_K}x_k - |\mathcal{S}_T| +\epsilon)$. The disjunctions in the SDNF are implemented in the RBM simply by creating a hidden neuron $h_j$ for each disjunct in the SDNF.

Learning in LBM uses learning from data $\mathcal{D}$ combined with knowledge provided by the logical formulae. Learning with data and knowledge is expected to improve accuracy or training time. If the logical formula is empty, the weights and biases are initialized randomly and one has a standard RBM. Learning in this case is an approximation of parameters $\Theta$ over a set of preferred models $\mathcal{D}=\{\vt{x}^{(n)}|
n=1,..,N\}$ of an unknown formula $\varphi^*$. Consider the case where the data set $\mathcal{D}$ is complete, i.e. it contains all
preferred models of an unknown $\varphi^*$. We will show that learning an RBM to represent the
SDNF of $\varphi^*$ is possible. Consider the gradient of the negative log-likelihood ($-\ell$) of an RBM:

{\footnotesize 
\begin{equation}
\frac{\partial {-}\ell}{\partial \Theta} =  \mathbf{E}[\frac{\partial \En(\vt{x},\vt{h})}{\partial \Theta}]_{\vt{h}|\vt{x} \in \mathcal{D}}  - \mathbf{E}[\frac{\partial \En(\vt{x},\vt{h})}{\partial \Theta}]_{\vt{h},\vt{x}}
\end{equation}
}

\noindent where $\mathbf{E}$ denotes the expected value. This function is not convex. Therefore, the RBM may not always converge to $\varphi^*$. Consider now the case where $\mathcal{D}$ is incomplete. At a local minimum, we have that
$\frac{\partial \text{-}\ell}{\partial w_{ij}} = -\frac{1}{N}\sum_{\vt{x} \in \mathcal{D}} x_ip(h_j|\vt{x}) + \sum_{ \vt{x}}x_ip(h_j|\vt{x})p(\vt{x}) \approx 0$.
A solution to this is
 $p(h_j|\vt{x})p(\vt{x}) \approx \frac{p(h_j|\vt{x})}{N} \text{ if } \vt{x} \in \mathcal{D}, \text{ and 0 otherwise.}$
This can be achieved by either having
$p(h_j|\vt{x})\approx 0$ or $p(\vt{x})\approx 0$ for all
$\vt{x} \notin \mathcal{D}$ and $p(\vt{x}) \approx \frac{1}{N}$ for
$\vt{x} \in \mathcal{D}$. Since $p(\vt{x})
= \frac{1}{Z}\sum_\vt{h} \exp(-\En(\vt{x},\vt{h}))$ then for a
training example (preferred model) $\vt{x}$ we have
$\sum_{\vt{x}} \sum_\vt{h} \exp(-\En(\vt{x},\vt{h})) \approx N \sum_\vt{h} \exp(-\En(\vt{x},\vt{h})) 
$
. Hence, a solution is obtained if
$\sum_\vt{h}\exp(-\En(\vt{x},\vt{h}))$ is equally large for all
$x \in \mathcal{D}$, and much smaller otherwise. We can further
factorize this sum to get
$\sum_\vt{h}\exp(-\En(\vt{x},\vt{h})) \propto \prod_j(1+\exp(\sum_i
w_{ij}x_i + \theta_j))$. Now, suppose that an LBM with parameters $\Theta=(W^*,\theta^*)$ represents an unknown
formula $\varphi^*$. 
Assuming that the LBM has large and equal confidence values
$c_{\infty}$ for its free-energy function (as discussed in Section \ref{cond_dis}), this LBM would allow only one hidden unit to be
activated for a satisfying assignment $\vt{x}$. In the case of an unsatisfiable assignment, all hidden units would be deactivated. Therefore, one can choose
$c_{\infty}$ large enough to guarantee that a solution is found
because
$ \prod_j(1+\exp(\sum_iw^*_{ij}x_i + b^*_j)) \approx 
\exp(c_{\infty}\epsilon) \text{ if } \vt{x} \in \mathcal{D}. $

\begin{example} 
\label{exp:xor2en}
 We use the symbol $\xor$ to denote exclusive-or, that is $\pr{x} \xor \pr{y} \equiv ((\pr{x} \wedge \neg \pr{y}) \vee (\neg \pr{x} \wedge \pr{y}))$. The formula $\varphi \equiv (\pr{x} \xor \pr{y}) \fziff \pr{z}$ can be converted into the SDNF:

\begin{equation*}
\varphi \equiv (\neg \pr{x} \fzand \neg \pr{y} \fzand \neg \pr{z}) \fzor (\neg \pr{x}
\fzand \pr{y} \fzand \pr{z}) \fzor (\pr{x} \fzand \neg \pr{y} \fzand \pr{z}) \fzor (\pr{x} \fzand \pr{y}
\fzand \neg \pr{z})
\end{equation*}

For each conjunctive clause in $\varphi$, a corresponding term is added to the energy function. An RBM for the XOR formula $\varphi$ can be built as shown in Figure \ref{dnf_xor} for a choice of $\epsilon = 0.5$ and zero bias for the visible units ($\theta_i=0$). The energy function of this RBM is:

{
\begin{equation*} 
  \begin{aligned} \En &= -h_1(-x - y - z + 0.5) - h_2(x + y -  z - 1.5)
   - \\ & h_3(x+y-z-1.5) - h_4( -x + y + z - 1.5)
  \end{aligned}
\end{equation*}
}

\begin{figure}[ht]
\centering
\includegraphics[width=.6\textwidth]{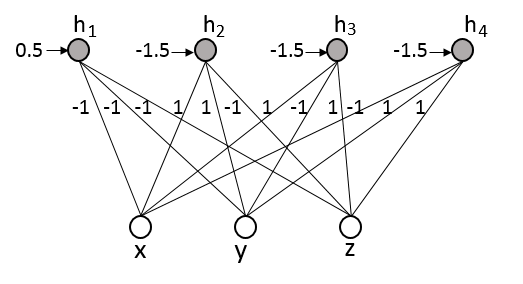}
\caption{An RBM equivalent to the XOR formula $(\pr{x} \xor \pr{y}) \fziff \pr{z}$.}
\label{dnf_xor}
\end{figure}

Table \ref{xor_example} shows the equivalence between $min_{\vt{h}}\En(\vt{x},\vt{h})$ and the truth-table for the XOR formula. The above example illustrates in a simple case the value of using SDNF, in that it produces a direct translation into an RBM, as also illustrated below. 

\begin{table}
\centering
\caption{Energy function and truth-table for the formula $((\pr{x} \wedge \neg \pr{y}) \vee (\neg \pr{x} \wedge \pr{y})) \fziff \pr{z}$.\label{xor_example}}
{%
\begin{tabular}{@{}ccc|cc@{}}
\toprule
$\pr{x}$     &   $\pr{y}$   & $\pr{z}$     & $s_\varphi(\pr{x},\pr{y},\pr{z})$  &  $min_{\vt{h}}\En(\vt{x},\vt{h})$ \\
\midrule
      $0$ & $0$ & $0$ & $True$              &  $-0.5$\\
      $0$ & $0$ & $1$ & $False$             &  $0$\\
      $0$ & $1$ & $0$ & $False$             &  $0$\\
      $0$ & $1$ & $1$ & $True$              &  $-0.5$\\
      $1$ & $0$ & $0$ & $False$             &  $0$\\
      $1$ & $0$ & $1$  & $True$              &  $-0.5$\\
      $1$ & $1$ & $0$ & $True$              &  $-0.5$\\
      $1$ & $1$ & $1$  &$False$             &  $0$\\
\toprule
\end{tabular}
}
\end{table}
\end{example}

 \begin{example}
 We have seen that the SDNF of $(\pr{x}_1 \fzand \pr{x}_2 \fzand \neg \pr{x}_3) \rightarrow \pr{y} $ is $\text{ } (\pr{y} \fzand \pr{x}_1 \fzand \pr{x}_2 \fzand \neg \pr{x}_3) \fzor (\pr{x}_1 \fzand \pr{x}_2 \fzand \pr{x}_3) \fzor (\pr{x}_1 \fzand \neg \pr{x}_2) \fzor \neg \pr{x}_1 $. We need an RBM with only 3 hidden units\footnote{In the case of $\neg x_1$, or any term of the energy function with a single variable, the term is implemented in the RBM via the bias of $x_1$. For a positive literal $x$, the energy term $-h(x-1+\epsilon)$ can be replaced by $-x\epsilon$. For a negative literal $\neg x$, the energy term $-h(-x+\epsilon)$ can be replaced by $-(1-x)\epsilon$. This is possible because in order to minimize the energy, $h=1$ if and only if $x=0$ (in the case of $\neg x$), thus $h=1-x$. Therefore, $-h(-x+\epsilon) = -(1-x)(-x+\epsilon)=-(-x +\epsilon+x^2-x\epsilon) = -(1-x)\epsilon$, because $x=x^2$.} to represent this SDNF. The energy function with $\epsilon=0.5$ is:
 
\begin{equation*}
\begin{aligned}
E= e_y + \sum_{i=1}^3 e_i, \text{where}\\
e_y = -h_y(y+x_1+x_2-x_3-2.5) & \quad \quad (\pr{y} \fzand \pr{x}_1 \fzand \pr{x}_2 \fzand \neg \pr{x}_3) \\
e_3 = -h_3(x_1 +x_2 + x_3 - 2.5) & \quad \quad (\pr{x}_1 \fzand \pr{x}_2 \fzand \pr{x}_3) \\
e_2 = -h_2(x_1-x_2-0.5) & \quad \quad (\pr{x}_1 \fzand \neg \pr{x}_2)\\
e_1=0.5x_1 & \quad \quad (\neg \pr{x}_1) 
\end{aligned}
\end{equation*}

 The number of hidden units grows linearly with the number of disjuncts in the formula. The computationally expensive part is the translation from WFF to SDNF in case it is needed.  
 \end{example}

\subsection{Experimental Results}
\label{exp}
\subsubsection{Reasoning} We deployed LBM to search for satisfying truth assignments of variables in large formulae. Let us define a class of formulae as:
\begin{equation}
    \varphi \equiv \bAnd_{i=1}^M \pr{x}_i \fzand (\bOr_{j=M+1}^{M+N} \pr{x}_j)
\end{equation}

A formula in this class consists of $2^{M+N}$ possible truth assignments of the variables, with $2^N-1$ of them mapping the formula to $true$ (call this the {\it satisfying set}). Converting to SDNF as done before but now for the class of formulae, we obtain:
\begin{equation}
    \varphi \equiv \bOr_{j=M+1}^{M+N} (\bAnd_{i=1}^M \pr{x}_i \fzand \bAnd_{j'=j+1}^{M+N} \neg \pr{x}_{j'} \fzand \pr{x}_{j} )
\end{equation}

\begin{figure}[ht]
\centering
\begin{subfigure}{0.45\textwidth}
\includegraphics[width=.99\textwidth]{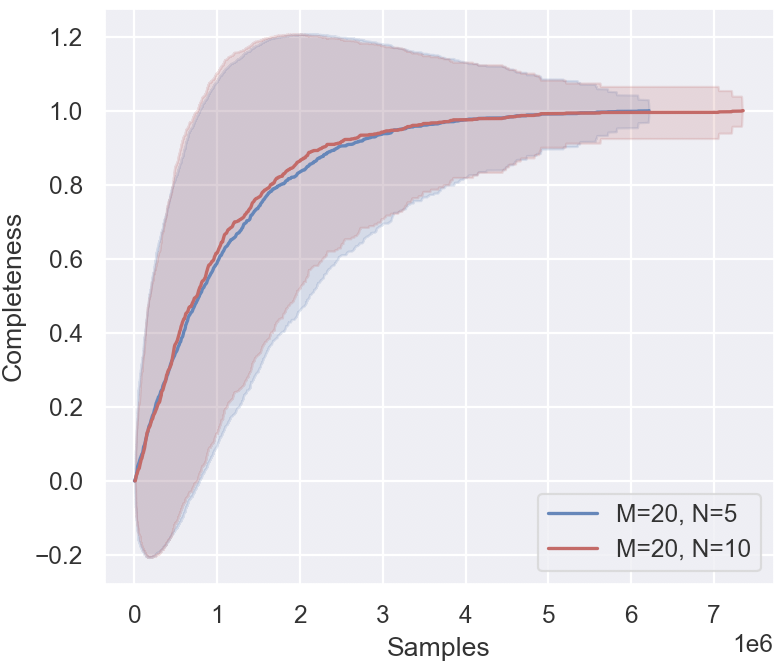}
\end{subfigure}
\centering
\begin{subfigure}{0.45\textwidth}
\includegraphics[width=.99\textwidth]{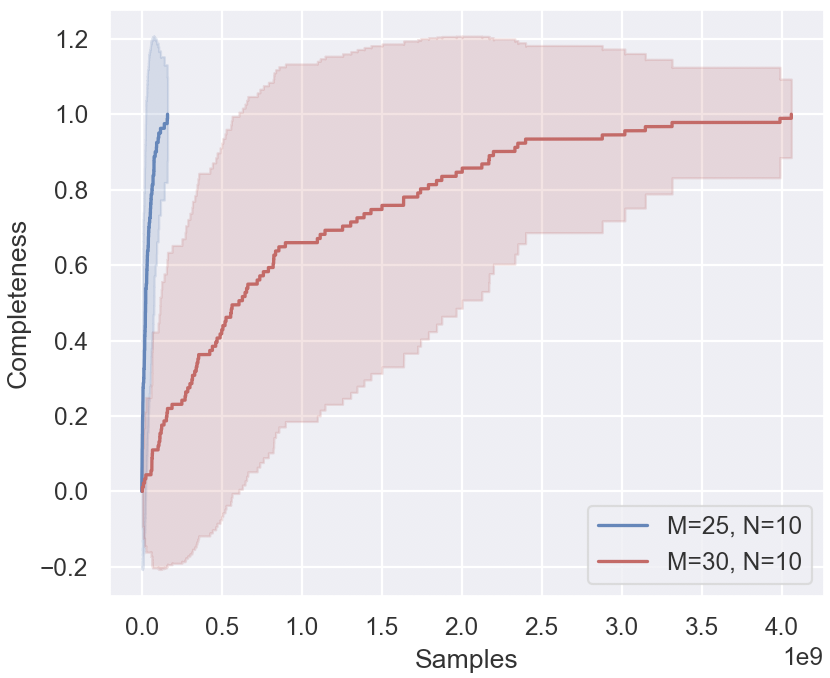}
\end{subfigure}
\caption{Percentage coverage as a measure of completeness as sampling progresses in the RBM. 100\% coverage is achieved for the class of formulae with different values for M and N averaged over 100 runs. The number of samples needed to achieve $100\%$ coverage is much lower than the number of possible assignments ($2^{M+N}$). For example, when M=20, N=10, all satisfying assignments are found after approximately $7.5 \times 10^6$ samples are provided as input to the RBM, whereas the number of possible assignments is approximately 1 billion, a ratio of sample size to the search space of $0.75\%$. The ratio for M=30, N=10 is even lower at $0.37\%$. Source: \cite{Tran_Garcez_2023}.}
\label{fig:sampling}
\end{figure}

\noindent Applying Theorem \ref{theorem:prop_rbm} to construct an RBM from $\varphi$, we use Gibbs sampling to find the models of a formula given random initial truth assignments to all the variables. A sample is {\it accepted} as a satisfying assignment (a model) if its free energy is lower than or equal to $-\log(1+\exp(c\epsilon)$ with $c=5, \epsilon=0.5$. We evaluate the {\it coverage} and {\it accuracy} of accepted samples. Coverage is measured as the proportion of the satisfying set that is accepted over time. In this experiment, this is the number of satisfying assignments in the set of accepted samples divided by $2^N-1$. It can be seen as a measure of completeness. Accuracy is measured as the percentage of samples accepted by the RBM that do satisfy the logical formula. 

We test different values of $M\in \{20,25,30\}$ and $N\in\{3,4,5,6,7,8,9,10\}$. LBM achieves $100\%$ accuracy in all cases, meaning that all accepted samples do satisfy the formula, as expected (given Theorem \ref{theorem:prop_rbm}).
Figure \ref{fig:sampling} shows the coverage as Gibbs sampling progresses (after each time that a number of random samples is collected). Four cases are considered: M=20 and N=5, M=20 and N=10, M=25 and N=10, M=30 and N=10. 

In each case, we run the sampling process 100 times and report the average results with standard deviations. The satisfying set and therefore the number of samples needed to achieve $100\%$ coverage is much lower than the number of possible assignments ($2^{M+N}$). For example, when M=20, N=10, all satisfying assignments are found after 7.5 million samples are collected, whereas the number of possible assignments is approximately 1 billion, producing a ratio of sample size to the search space size of just $0.75\%$. The ratio for M=30, N=10 is even lower at $0.37\%$ w.r.t. $10^{12}$ possible assignments. 

\begin{figure}[ht]
    \centering
    \includegraphics[width=0.6\textwidth]{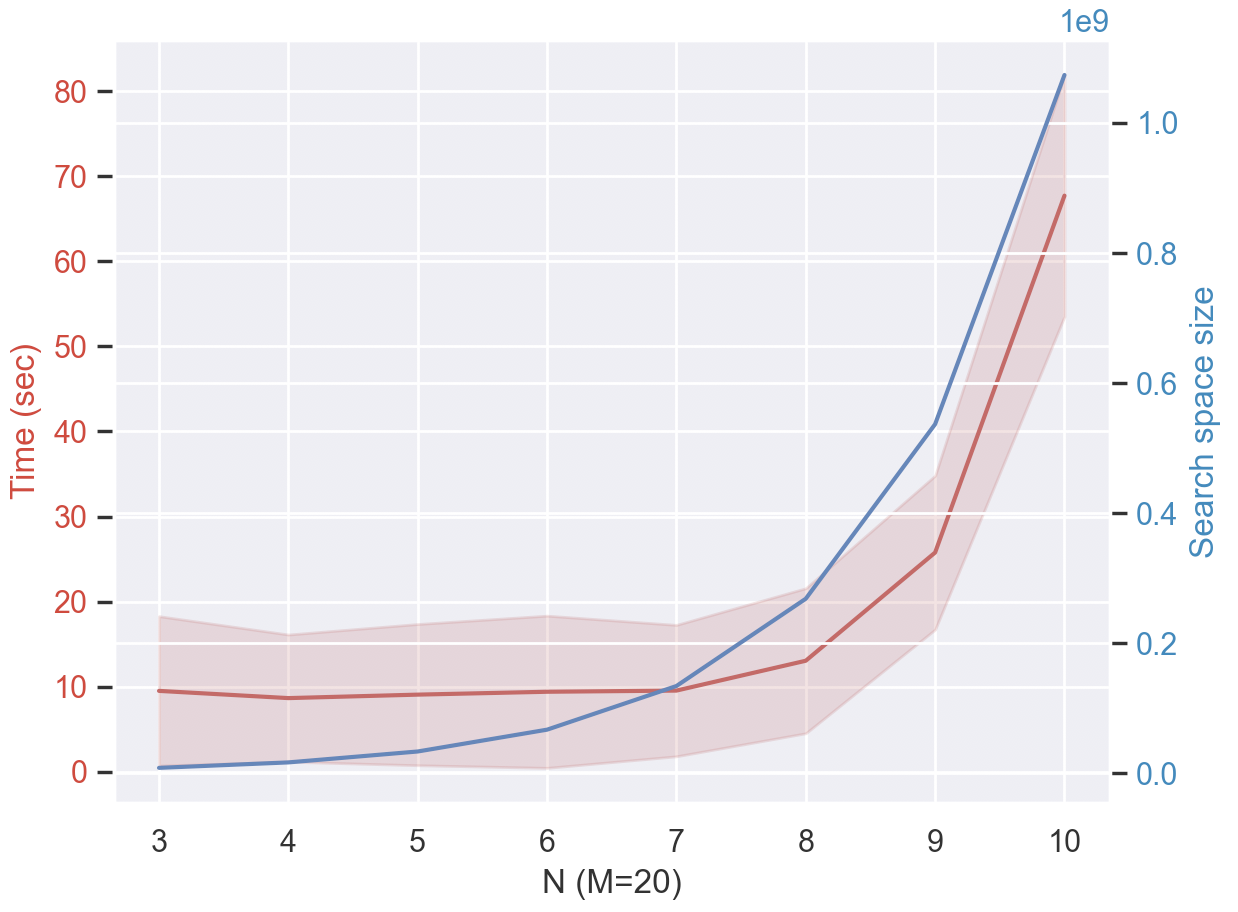}
    \vskip -.3cm
    \caption{Time taken by LBM to collect all satisfying assignments compared with the size of the search space (i.e. the number of possible assignments up to 1 billion (1e9)) as N increases from 3 to 10 with fixed M=20. LBM only needs around 10 seconds for $N<=8$, $\sim 25$ seconds for $N=9$, and $\sim 68$ seconds for $N=10$. The curve grows exponentially, similarly to the search space size, but at a much lower scale. Source: \cite{Tran_Garcez_2023}.}
    \label{fig:time}
\end{figure}

Figure \ref{fig:time} shows the time taken to collect all satisfying assignments for different N in $\{3,4,5,6,7,8,9,10\}$ with $M=20$. LBM needed around 10 seconds for $N<=8$, $25$ seconds for $N=9$, and $68$ seconds for $N=10$. As expected, the curve grows exponentially similarly to the search space curve, but at a much smaller scale. 


\subsubsection{Learning from Data and Knowledge} We now evaluate LBM at learning the same Inductive Logic Programming (ILP) benchmark tasks used by neurosymbolic system CILP++ \cite{Franca_2014} in comparison with ILP state-of-the-art system Aleph \cite{aleph}. As mentioned earlier, the systems Aleph, CILP++ and a fully-connected standard RBM were chosen as the natural symbolic, neurosymbolic and neural system, respectively, for comparison. An initial LBM is constructed from the clauses provided as background knowledge. This process creates one hidden neuron per clause. Further hidden neurons are added using random weights for training and validation from data. Satisfying assignments can be selected from each clause as a training or validation example, for instance given clause $\pr{x}_1 \fzand \neg \pr{x}_2 \rightarrow y$, assignment $x_1=True,x_2=False,y=True$ is converted into vector $[x_1,x_2,y]=(1,0,1)$ for training or validation. Both the LBM and the standard RBM are trained discriminatively using the conditional distribution $p(y|\vt{x})$ for inference as in \cite{Larochelle_2012}. In both cases, all network weights are free parameters for learning, with some weights having been initialized by the background knowledge in the case of the LBM, that is, the background knowledge can be revised during learning from data.  

Seven data sets with available data and background knowledge (BK) are used: Mutagenesis (examples of molecules tested for mutagenicity and BK provided in the form of rules describing relationships between atom bonds) \cite{Srinivasan_1994}, KRK (King-Rook versus King chess endgame with examples provided by the coordinates of the pieces on the board and BK in the form of row and column differences) \cite{Bain_1994}, UW-CSE (Entity-Relationship diagram with data about students, courses taken, professors, etc. and BK describing the relational structure) \cite{Richardson_2006}, and the Alzheimer's benchmark: Amine, Acetyl, Memory and Toxic (a set of examples for each of four properties of a drug design for Alzheimer's disease with BK describing bonds between the chemical structures) \cite{King_1995}. With the clauses converted into their equivalent set of preferred models in the form of vectors such as $[x_1,x_2,y]$ above, and combined with the available data, for the Mutagenesis and KRK tasks, $2.5\%$ of the data is used to build the initial LBM. For the larger data sets UW-CSE and Alzheimer's, $10\%$ of the data is used as BK. The remaining data are used for training and validation based on 10-fold cross validation for each data set, except for UW-CSE that uses 5 folds for the sake of comparison. The number of hidden units added to the LBM is chosen arbitrarily at $50$. The standard RBM without BK is given a higher degree of freedom with $100$ hidden units. 
Results are shown in Table \ref{tab:ilp}.\footnote{The results for Aleph and CILP++ are obtained from \cite{Franca_2014}.} It can be seen that LBM has the best performance in 5 out of 7 data sets. Some of the results of the LBM and RBM are comparable when the BK can be learned from the examples, as in the case of the Alzheimer's amine data set. In these cases, training the LBM is faster than the RBM. Aleph is better than all other models in the {\it alz-acetyl} data set. This task must rely more heavily on the correctness of the BK than the data. CILP++ however is considerably faster than Aleph and it can achieve comparable results. Although direct comparisons of running times are not possible to make between CILP++ and LBM, LBM's running times look promising.

\begin{table}
\caption{Cross-validation performance of LBM against purely-symbolic system Aleph, neurosymbolic system CILP++ and a standard RBM on 7 benchmark data sets for neurosymbolic AI. We run cross-validation on RBM and LBM 100 times and report the average results with $95\%$ confidence interval. Source: \cite{Tran_Garcez_2023}.\label{tab:ilp}}
{%
\begin{tabular}{@{}lcccc@{}}
\toprule
            & { Aleph}    & {CILP++} & RBM     & {LBM}  \\ 

\midrule
{\small Mutagenesis}       &  ${80.85}$ ($\pm 10.5$ )    & ${91.70}(\pm 5.84)$ & ${95.55}(\pm 1.36)$ & ${\mathbf{96.28}}(\pm 1.21)$ \\
 {\small KRK}        & ${99.60}(\pm 0.51)$    & ${98.42}(\pm 1.26 )$       & ${99.70}(\pm 0.11)$ & ${\mathbf{99.80}}(\pm 0.09)$ \\
 {\small UW-CSE}     & ${84.91}(\pm 7.32)$    & ${70.01}(\pm 2.2)$ & ${89.14}(\pm 0.46)$      & ${\textbf{89.43}}(\pm 0.42)$ \\
 {\small alz-amine}  & ${78.71}(\pm 5.25)$    & ${78.99}(\pm 4.46)$ & ${\textbf{79.13}}(\pm 1.14)$     & ${78.25}(\pm 1.07)$ \\
 {\small alz-acetyl} & ${\textbf{69.46}}(\pm 3.6)$    & ${65.47}(\pm 2.43)$        & ${62.93}(\pm 0.31)$ & ${66.82}(\pm 0.28)$ \\
 {\small alz-memory} & ${68.57}(\pm 5.7)$    & ${60.44}(\pm 4.11)$      & ${68.54}(\pm 0.97)$ & ${\mathbf{71.84}}(\pm 0.88)$ \\
 {\small alz-toxic}  & ${80.50}(\pm 3.98)$    & ${81.73}(\pm 4.68)$     & $82.71 (\pm 1.18)$  & $\mathbf{84.95} (\pm 1.04)$ \\
\toprule
\end{tabular}
}
\end{table}
  
 \subsection{Extensions of Logical Boltzmann Machines}

\subsubsection{Translating CNF into RBMs}
\label{supp:CNF}

In the general case, translation to SDNF can be costly. When knowledge is provided in CNF form, it is useful to be able to translate the CNF directly into the RBM without the need for an intermediate step.

Every WFF can be converted into CNF. A CNF is a conjunction of clauses. Formally:
\begin{equation}
    \varphi_{\text{CNF}} \equiv \bAnd_{m=1}^M(\bOr_{t\in\mathcal{S}^m_T} x_t \fzor  \bOr_{k\in\mathcal{S}^m_K} \neg x_k)
\end{equation}

We will apply the same transformation process into SDNF to each conjunctive clause in the CNF. The result will be a conjunction of $M$ SDNFs (itself not an SDNF), as follows:

\begin{equation}
\label{dnf_form}
\begin{aligned}
\varphi_{\text{CNF}} \equiv \bAnd_{m=1}^M (\bOr_{t\in \mathcal{S}^m_T} \neg \pr{x}_t \fzor \bOr_{k\in \mathcal{S}^m_K} \pr{x}_k) \equiv \bAnd_{m=1}^M (\bOr_{p \in \mathcal{S}^m_T \cup \mathcal{S}^m_K} (\bAnd_{t\in\mathcal{S}^m_T\backslash p} \pr{x}_t \fzand  \bAnd_{k\in\mathcal{S}^m_K\backslash p} \neg\pr{x}_k \fzand \pr{x}'_{p}))
\end{aligned}
\end{equation}

\noindent where $\pr{x}'_p\equiv\neg\pr{x}_p$ if $p\in\mathcal{S}^m_T$; otherwise $\pr{x}'_p\equiv\pr{x}_p$.

This transformation would increase the space complexity from $\mathcal{O}(M\times N)$ to $\mathcal{O}(M\times N^2)$, where $M$ is the number of clauses and $N$ is the number of variables. This should not be a problem for current computing systems, especially when inference with RBMs can be highly parallelized.

Although the formula in Eq.\eqref{dnf_form} is not a SDNF, the equivalence between the CNF and the LBM still holds:
\begin{equation}
\label{dnf_equiv}
s_\varphi = \begin{dcases*}
        1  & when $-\frac{1}{\epsilon}min_\vt{h} E(\vt{x},\vt{h})=M$\\
        0 & otherwise
        \end{dcases*}
\end{equation}

Eq.\eqref{dnf_equiv} holds because the CNF is satisfied if and only if all $M$ SDNFs are satisfied. Under such circumstances, $min_\vt{h}E(\vt{x},\vt{h})=-M\epsilon$. Otherwise, $min_\vt{h}E(\vt{x},\vt{h})=-M'\epsilon$, where $M'<M$.

When a {\it confidence value} c is used, the number of satisfied clauses in CNF will be proportional to the minimized energy function, and to the free-energy function when c increases. 

\subsubsection{Towards using LBM as a SAT Solver}
\label{supp:SAT}
The Boolean satisfiability (SAT) problem is a fundamental problem in Computer Science. It was the first problem that was proven to be NP complete. A formula is satisfiable if and only if there exists an assignment of truth-values mapping the formula to True. In practice, formulae in SAT problems are represented as Conjunctive Normal Forms (CNFs).

As discussed in Section \ref{supp:CNF}, a formula in Conjunctive Normal Form (CNF) can be converted into a Logical Boltzmann Machine (LBM). The number of satisfied clauses in the CNF formula is proportional to the minimized energy function and the free-energy function of the LBM. This relationship allows us to solve SAT problems by transforming them into an optimization task: finding the minimum of the energy or free-energy function.

\begin{figure}[ht]
    \centering
    \begin{subfigure}{0.24\textwidth}
        \includegraphics[width=1.1\textwidth]{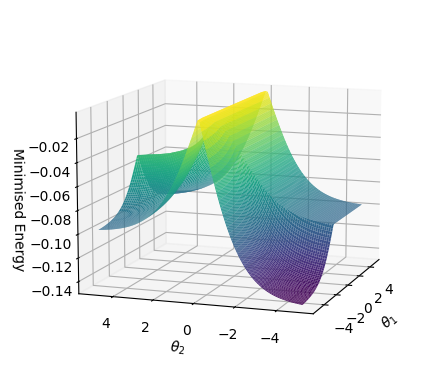}
        \caption{c=0.1}
        \label{ec_0.1}
    \end{subfigure}
    \begin{subfigure}{0.24\textwidth}
        \includegraphics[width=1.1\textwidth]{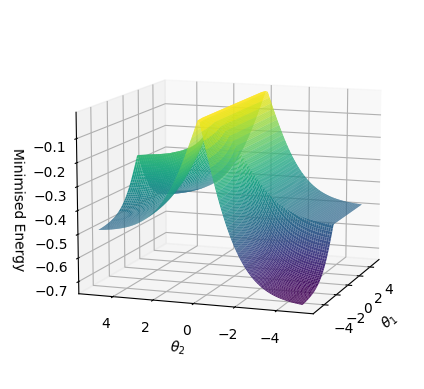}
        \caption{c=0.5}
        \label{ec_0.5}
    \end{subfigure}
    \begin{subfigure}{0.24\textwidth}
        \includegraphics[width=1.1\textwidth]{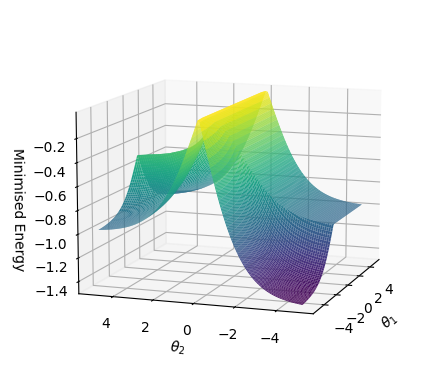}
        \caption{c=1}
        \label{ec_1.0}
    \end{subfigure}
    \begin{subfigure}{0.24\textwidth}
        \includegraphics[width=1.1\textwidth]{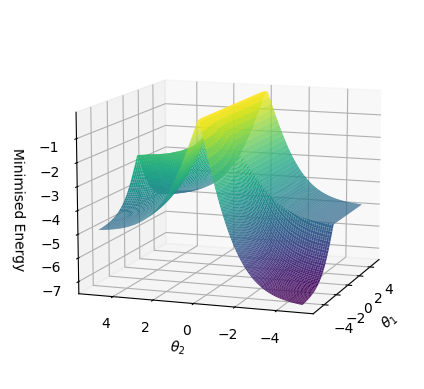}
        \caption{c=5}
        \label{ec_5.0}
    \end{subfigure}
    \begin{subfigure}{0.24\textwidth}
        \includegraphics[width=1.1\textwidth]{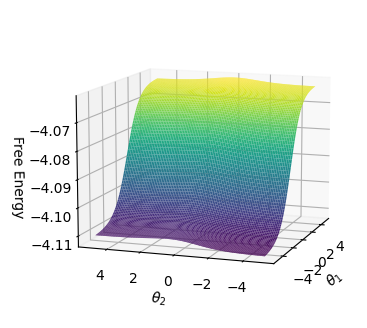}
        \caption{c=0.1}
        \label{fc_0.1}
    \end{subfigure}
    \begin{subfigure}{0.24\textwidth}
        \includegraphics[width=1.1\textwidth]{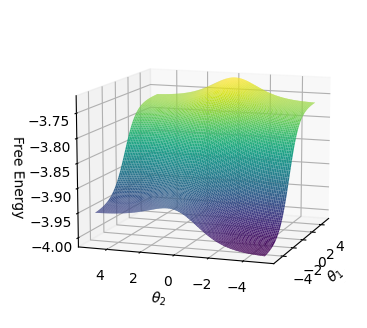}
        \caption{c=0.5}
        \label{fc_0.5}
    \end{subfigure}
    \begin{subfigure}{0.24\textwidth}
        \includegraphics[width=1.1\textwidth]{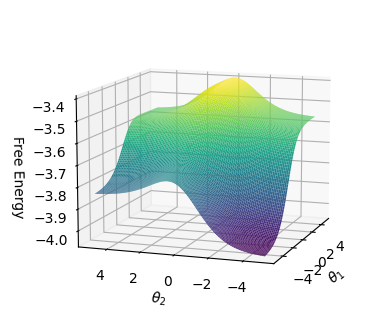}
        \caption{c=1}
        \label{fc_1.0}
    \end{subfigure}
    \begin{subfigure}{0.24\textwidth}
        \includegraphics[width=1.1\textwidth]{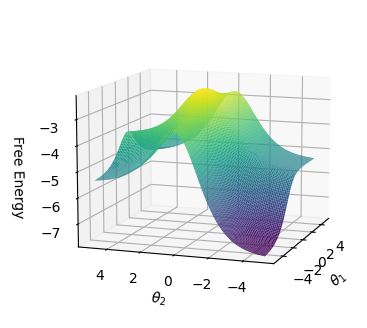}
        \caption{c=5}
        \label{fc_5.0}
    \end{subfigure}
    \caption{Energy function and free-energy function with different {\it confidence values}.}
    \label{fig:sat_exp}
\end{figure}

To make this approach computationally feasible, we focus on minimizing the free-energy function, as it is both easier to compute and differentiable. This transformation converts the discrete SAT problem into a continuous optimization problem. Instead of searching for solutions in a Boolean space (where variables $ x $ are either 0 or 1), we search in a continuous space for parameters $ \theta $, where each Boolean variable $ x $ is represented as a sigmoid function:
$$
x = \sigma(\theta) = \frac{1}{1 + \exp(-\theta)}.
$$
This mapping ensures that $ x $ smoothly transitions between 0 and 1 as $ \theta $ changes, enabling gradient-based optimization techniques to be applied.
To illustrate this process, consider a simple SAT problem with two variables:
$$
(\neg x_1 \lor \neg x_2) \land (x_1 \lor \neg x_2) \land (\neg x_1 \lor x_2).
$$
Figure \ref{fig:sat_exp} visualizes the landscape of the LBM's energy and free-energy functions for different values of $ \theta_1 $ and $ \theta_2 $, where $ x_1 = \sigma(\theta_1) $ and $ x_2 = \sigma(\theta_2) $. The plots reveal that when both $ \theta_1 $ and $ \theta_2 $ are more negative (corresponding to $ x_1, x_2 \approx 0 $), the functions approach their minima. This corresponds to a satisfying assignment of the CNF formula, illustrating how the optimization process identifies valid solutions.

We also analyze the impact of confidence values $ c $ on the landscapes of the energy and free-energy functions. Figures \ref{ec_0.1}, \ref{ec_0.5}, \ref{ec_1.0}, and \ref{ec_5.0} show that confidence values do not significantly alter the landscape of the energy function. However, for the free-energy function (Figures \ref{fc_0.1}, \ref{fc_0.5}, \ref{fc_1.0}, \ref{fc_5.0}), smaller values of $ c $ result in smoother landscapes. While this smoothing effect can facilitate optimization by reducing sharp transitions, it also narrows the gap between local minima and the global minimum. Conversely, higher values of $ c $ increase the boundaries between optimal regions, making it more challenging to locate the global optimum. This trade-off highlights the importance of carefully selecting $ c $ based on the specific characteristics of the SAT problem being solved.

In summary, the LBM framework provides an approach to solving SAT problems by converting them into continuous optimization tasks. By leveraging the differentiability of the free-energy function and the flexibility of sigmoid mappings, this approach bridges logical reasoning and numerical optimization. Future work should explore adaptive strategies for adjusting confidence values to balance smoothness and optimality and consider ways to enhance performance in the case of specific classes of SAT problems. 

\subsubsection{Comparison to Other Approaches} Unlike recent neural network-based approaches for SAT solving \cite{selsam2018learning, satnet}, which rely on large datasets generated by traditional SAT solvers for training, our method eliminates this dependency. Instead, we directly convert SAT problems (typically expressed in CNF) into LBM representations. This data-independent transformation should simplify the workflow by reducing the need for extensive pre-processing or model training. When compared with other SAT-solving methods that utilize Boltzmann Machines (BM), such as those in \cite{SAT1, SAT2}, our approach leverages the much simpler structure of Restricted Boltzmann Machines (RBMs). These earlier methods often employ dense or higher-order structures, which are computationally complex and challenging to implement. By contrast, our use of RBMs maintains a streamlined architecture without requiring modifications to the network (e.g. adding configurations) to map SAT problems to BMs. Finally, while LBM is not yet competitive with state-of-the-art SAT solvers in terms of raw performance, it seems to open up a promising direction for further research. Unlike traditional SAT solvers, LBM is in essence a neural network. This should enable a seamless integration of reasoning and learning. Furthermore, LBM does not require prior knowledge of SAT problem structures or specific solving tactics such as backtracking or unit propagation. It is important to notice that our current implementation relies solely on off-the-shelf optimization methods, with room for significant improvements through parallelization optimization and the incorporation of advanced SAT-solving strategies. 

\subsubsection{SAT Solving Methodology and Initial Experimental Results} We used random SAT problems \cite{Amizadeh2019LearningTS} as a case study. To solve SAT problems using LBM, we follow these steps:
\begin{itemize}
    \item Convert a CNF into an RBM using the transformation described earlier.
    \item Apply various inference and optimization techniques to find satisfying assignments or determine unsatisfiability. Specifically:
    \begin{itemize}
        \item Use Gibbs sampling to minimize the energy function and search for satisfying assignments.
        \item Employ gradient-based methods from TensorFlow to optimize the differentiable free energy function.
        \item Utilize stochastic optimization methods from the Scipy library, such as {\it dual\_annealing} and {\it differential\_evolution}.
    \end{itemize}
\end{itemize}

Our experiments produced the following results:
\begin{itemize}
    \item \textbf{Gibbs Sampling:}
    \begin{itemize}
        \item Gibbs sampling can find satisfying assignments for CNFs with fewer than 40 variables. For larger formulas, Gibbs sampling often gets stuck in local minima, making it challenging to determine satisfiability. In such cases, if the free energy function does not decrease after 1000 steps of Gibbs sampling, we conclude that the formula is likely unsatisfiable.
    \end{itemize}

    \item \textbf{Gradient-Based Optimization (TensorFlow):}
    \begin{itemize}
        \item Gradient-based methods are applied to minimize the differentiable free energy function. However, these methods are prone to getting trapped in local minima, especially for SAT problems with more than 20 variables.
        \item Attempts to smoothing the energy landscape by adjusting confidence values did not significantly improve performance, highlighting a difficulty in solving larger SAT instances with this approach.
    \end{itemize}

    \item \textbf{Stochastic Optimization (Scipy):}
    \begin{itemize}
        \item Among the stochastic optimization methods available in Scipy, {\it dual\_annealing} and {\it differential\_evolution} showed better scalability. These methods successfully solved SAT problems with up to 100 variables.
        \item This suggests that stochastic global optimization techniques may offer a viable alternative for solving larger SAT instances with RBMs.
    \end{itemize}
\end{itemize}


{


{


 

\subsubsection{Implementing Penalty Logic in LBM}
\label{supp:PL}
\label{vspenalty}
The closest work to LBM is Penalty Logic \cite{Pinkas_1995}, which represents propositional formulae in Hopfield networks and Boltzmann machines. In its first step, Penalty Logic creates \textit{hidden variables} to reduce a formula $\varphi$ to a conjunction of sub-formulas $\bAnd_i \varphi_i$, each with at most three variables. This \textit{naming} step makes conversion into an energy function easier, but some of the terms in the energy function may consist of hidden variables and therefore cannot be converted into an RBM. For example, a negative term $-h_1xy$ of a higher-order Boltzmann Machine would be transformed into the quadratic term $-h_2h_1 - h_2x -h_2y + 5h_2$ with $-h_2h_1$ forming a connection between two hidden units, which is not allowed in RBMs. The three-variable term $-h_1xy$ is implemented in the higher-order network as a hypergraph. The variable $h_2$ is introduced to turn the hyper-edge into normal edges between each of the three variables and $h_2$ with an appropriate bias value for the new node $h_2$, in this example a value of 5.\footnote{Contrast the LBM for XOR in Figure \ref{dnf_xor} with the RBM for XOR built using Penalty Logic: first, one computes the higher-order energy function:
$
\label{eq:xor_en}
\En^p =  4xyz - 2xy - 2xz - 2yz + x +y +z,
$
then transforms it to quadratic form by adding a hidden variable $h_1$ to obtain:
$
\En^p = 2xy - 2xz - 2yz - 8xh_1 - 8yh_1 + 8zh_1 + x + y + z + 12h_1,
$
which is not an energy function of an RBM, so one keeps adding hidden variables until the energy function of an RBM might be obtained, in this case: $
\En^p = -8xh_1 - 8yh_1 + 8zh_1 + 12h_1 
    -4xh_2 + 4yh_2 + 2h_2 - 4yh_3 - 4zh_3 + 6h_3 -4xh_4 - 4zh_4 + 6h_4  + 3x + y + z.
$} 

The LBM system converts any set of formulae $\Phi = \{\varphi_1, ...,\varphi_n\}$ into an RBM by applying Theorem 1 to each formula
$\varphi_i \in \Phi$. In the case of Penalty Logic, formulae are weighted. Given a set of weighted
formulae $\Phi = \{w_1: \varphi_1, ..., w_n:\varphi_n\}$, one can also construct an equivalent RBM where each energy term generated from formula $\varphi_i$ is multiplied by $w_i$. In both cases, the assignments that minimise the energy of the RBM are the assignments that maximise the satifiability of $\Phi$, i.e. the (weighted) sum of the truth-values of the formula.  

\begin{lemma}
  \label{prop:rbm_lprogram} Given a weighted knowledge-base
$\Phi=\{w_1:\varphi_1,...,w_n:\varphi_n\}$, there exists an
equivalent RBM $\mathcal{N}$ such that
$s_\Phi(\vt{x}) = -\frac{1}{\epsilon}min_{\vt{h}}\En(\vt{x},\vt{h})$, where $s_\Phi(\vt{x})$ is the sum of the weights of the formulae in $\Phi$ that are satisfied by assignment $\vt{x}$.
\end{lemma}

A formula $\varphi_i$ can be decomposed into a set of (weighted) conjunctive clauses from its SDNF. If there exist two conjunctive clauses such that one is subsumed by the other then the subsumed clause is removed and the weight of the remaining clause is replaced by the sum of their weights. Identical conjunctive clauses are treated in the same way: one of them is removed and the weights are added. 
From Theorem 1, we know that a conjunctive clause
$\bAnd_{t \in \mathcal{S}_{T_j}}\pr{x}_t \fzand \bAnd_{k \in
\mathcal{S}_{K_j}} \neg \pr{x}_{k}$ is equivalent to an energy term
$e_j(\vt{x},h_j) = -h_j(\sum_{t\in \mathcal{S}_{T_j}} x_t
- \sum_{k \in \mathcal{S}_{K_j}}x_{k} -|\mathcal{S}_{T_{j}}|+ \epsilon)$ where $0<\epsilon<1$. A weighted conjunctive clause $w': \bAnd_{t \in \mathcal{S}_{T_j}}\pr{x}_t \fzand \bAnd_{k \in
\mathcal{S}_{K_j}} \neg \pr{x}_{k}$, therefore, is equivalent to an energy term
 $w' e_j(\vt{x},h_j)$. For each weighted conjunctive clause, we can add a hidden unit $j$ to an RBM with connection weights $w_{tj} = w'$ for all $t\in \mathcal{S}_{T_j}$, and $w_{kj}=-w'$ for all $k \in
 \mathcal{S}_{K_j}$. The bias for this hidden unit will be $w'(-|\mathcal{S}_{T_{j}}|+\epsilon)$. The weighted knowledge-base and the RBM are equivalent because $s_\Phi(\vt{x})
 \propto -\frac{1}{\epsilon} min_{\vt{h}}\En(\vt{x},\vt{h})$, where $s_\Phi(\vt{x})$ is the
 sum of the weights of the clauses that are satisfied by $\vt{x}$.

\begin{example} \label{nixonexample} (Nixon diamond problem) Consider the following weighted knowledge-base from the original Penalty Logic paper \cite{Pinkas_1995} (the weights of 1000 and 10 are given and have been taken from the original paper):
{
\begin{align*}
&1000: \pr{n} \rif \pr{r}  \quad \text{ Nixon is a Republican.} \\       
&1000: \pr{n} \rif \pr{q}  \quad \text{ Nixon is also a Quaker.}\\
&10\text{ }\text{ }\text{ }  : \pr{r} \rif \neg \pr{p} \quad \text{Republicans tend not to be Pacifists.}\\
&10\text{ }\text{ }\text{ }  : \pr{q} \rif \pr{p} \quad \text{Quakers tend to be Pacifists.}
\end{align*}
}
\begin{figure}[ht]
\centering
\includegraphics[width=0.45\textwidth]{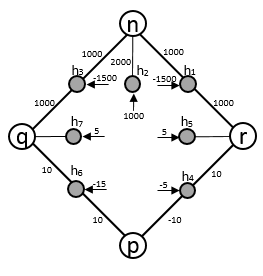}
\caption{The RBM for the Nixon diamond problem has 4 input neurons $\{n,q,r,p\}$ and 7 hidden neurons (shown in grey) as a result of the conversion into SDNF of the 4 weighted clauses shown in Example \ref{nixonexample}. }
\label{fig:rbm_nixon}
\end{figure}

Converting all four weighted clauses above into SDNF produces eight conjunctive clauses. For example, weighted clause $1000:\pr{n} \rif \pr{r} \equiv 1000:(\pr{n} \fzand \pr{r})\fzor (\neg \pr{n})$. After adding the weights of clause ($\neg \pr{n}$) which appears twice, an RBM is created (Figure \ref{fig:rbm_nixon}) representing the following unique conjunctive clauses with their corresponding confidence values:
$ 1000: \pr{n} \fzand \pr{r}, \quad 2000: \neg \pr{n}, \quad 1000: \pr{n} \fzand \pr{q}, \quad 10: \pr{r} \fzand \neg \pr{p}, \quad 10: \neg \pr{r}, \quad 10: \pr{q} \fzand \pr{p}, \quad 10: \neg \pr{q}.
$

With $\epsilon=0.5$, this RBM has energy function:
$
\En = -h_1(1000n+1000r-1500) - h_2(-2000n+1000)-h_3(1000n + 1000q-1500)-h_4(10r - 10p - 5)-h_5(-10r + 5) -h_6(10q + 10p - 15)- h_7(-10q + 5).
$
\end{example}



{

%% file: edjard2.tex


\label{sec:ms}
MaxSAT - shorthand for Maximum Satisfiability - is a computational problem that extends the classical SAT (Boolean satisfiability) problem. In MaxSAT, the goal is to find an assignment of truth values to the variables of a Boolean formula that maximizes the number of satisfied clauses. The formula is typically represented in CNF. We denote the number of satisfied clauses given a assignment $\vt{x}$ as $\sum_m \mathbbm{1}(\vt{x}\models\varphi_m)$. Here, $\vt{x}\models\varphi_m$ denotes that an assignment $\vt{x}$ satisfies the clause $\varphi_m$ of the CNF and $\mathbbm{1}$ is an indicator function mapping a clause-satisfying assignment to 1, and everything else to 0. Differently from SAT, where the goal is to find any satisfying assignment, MaxSAT seeks the assignment that satisfies the maximum number of clauses, making it a combinatorial optimization problem:

\begin{equation}
    \vt{x}^* = \argmax_\vt{x}\sum_m \mathbbm{1}(\vt{x}\models\varphi_m)
\end{equation}


For example, consider the CNF formula:
\[
\phi = (\neg x_1 \lor \neg x_2) \land (\neg x_1 \lor x_2) \land (x_1 \lor \neg x_2) \land (x_1 \lor x_2). 
\]

The goal is to find an assignment $ x \in \{0, 1\}^n $ (where $ n $ is the number of variables, in this example $n=2$) that maximizes the number of satisfied clauses. We know that $\phi$ is unsatisfiable since each clause corresponds to exactly one assignment. An optimal solution will satisfy three of the four clauses. This relaxation of the SAT task makes MaxSAT particularly suited for real-world optimization problems where constraints may need to be prioritized or relaxed to achieve the best overall solution. However, this flexibility comes at a cost: NP-Hardness, meaning that finding exact solutions becomes computationally infeasible as the number of variables grows. Many real-world applications, however, require approximate solutions to the MaxSAT problem, with the main challenge being how to balance accuracy and computation time.

Symbolic MaxSAT solvers have been the focus of intensive research for many years, gaining popularity among researchers and finding application in various domains from AI and computer-aided design to automated reasoning. Recent advancements in MaxSAT solvers have demonstrated significant improvements, with state-of-the-art solvers capable of scaling up to handling millions of variables and clauses. 

MaxSAT has proven to be valuable in software analysis \cite{DBLP:conf/cav/SiZGN17}, hardware verification \cite{maxsat_hardw,maxsat_hardw1}, combinatorial optimization \cite{maxsat_comb}, bioinformatics \cite{maxsatbio}, and data analysis \cite{POS-18:Applications_of_MaxSAT_in}. Despite these achievements, the growing complexity of real-world problems has spurred interest in alternative approaches, such as leveraging the differentiable capabilities of Machine Learning to address MaxSAT by transforming the discrete problem into a continuous optimization task.

In recent years, there has been a growing interest in connectionist solvers. These solvers aim to represent MaxSAT problems using neural networks to benefit from the parallel computation and learning capabilities of such ML systems and from specialized neural network accelerators, such as GPUs and TPUs, to further enhance performance. Beyond providing an alternative approach to solving complex reasoning problems, connectionist MaxSAT solvers may promote the development of interpretable classification models \cite{MM18}, facilitating novel neurosymbolic learning and reasoning \cite{satnet, Tran_Garcez_2023} with the promise of offering more transparent decision making in AI. 

Deep learning-inspired MaxSAT solvers rely on real-valued weights often learned from example solutions \cite{satnet}. In a related attempt, \cite{gnn_maxsat} seeks to train a Graph Neural Network for MaxSAT problem-solving. Unlike symbolic approaches, ML-based methods typically require a degree of supervision and may be criticized for their opacity. We take a different approach and use LBM along with a global optimization method called dual annealing, a modified version of a meta-heuristic method known as simulated annealing, to search for MaxSAT solutions. Using dual annealing, we will search the energy landscape of the RBM for a global minimum corresponding to a MaxSAT solution.

Research that has also focused on representing MaxSAT problems in connectionist networks without relying on explicit learning from examples include \cite{10.1007/978-1-4471-2063-6_120}, where the MaxSAT problem is mapped to a combinatorial optimization framework based on a high-order Boltzmann Machine used to search for an approximate solution to SAT. However, the problems with efficiency of Boltzmann Machines are well-known. They have motivated the use of RBMs, first in \cite{Tran_Garcez_2023} and leading up to this work. In a more recent development also using RBMs, RBMSAT was proposed to construct an RBM that represents the probability of an assignment w.r.t. the number of clauses that it satisfies \cite{rbmsat}. The goal there is to conduct a heuristic search for solutions using block Gibbs samplings on neural network accelerators. Our goal, instead, is to make use of an interpretable RBM layer, as shown e.g. in Figure \ref{fig:sii} where each clause can be read off the LBM with binary weights.

\begin{figure*}[ht]
    \begin{subfigure}{0.3\textwidth}
    \includegraphics[width=0.9\textwidth]{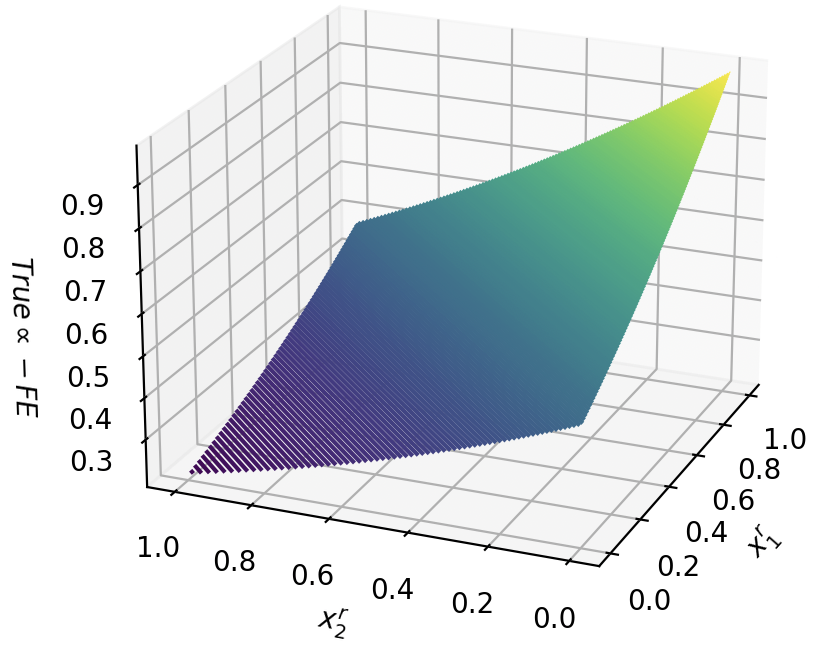}
    \caption{c=1}
    \label{fig:c1}
    \end{subfigure}
    \begin{subfigure}{0.3\textwidth}
    \includegraphics[width=0.9\textwidth]{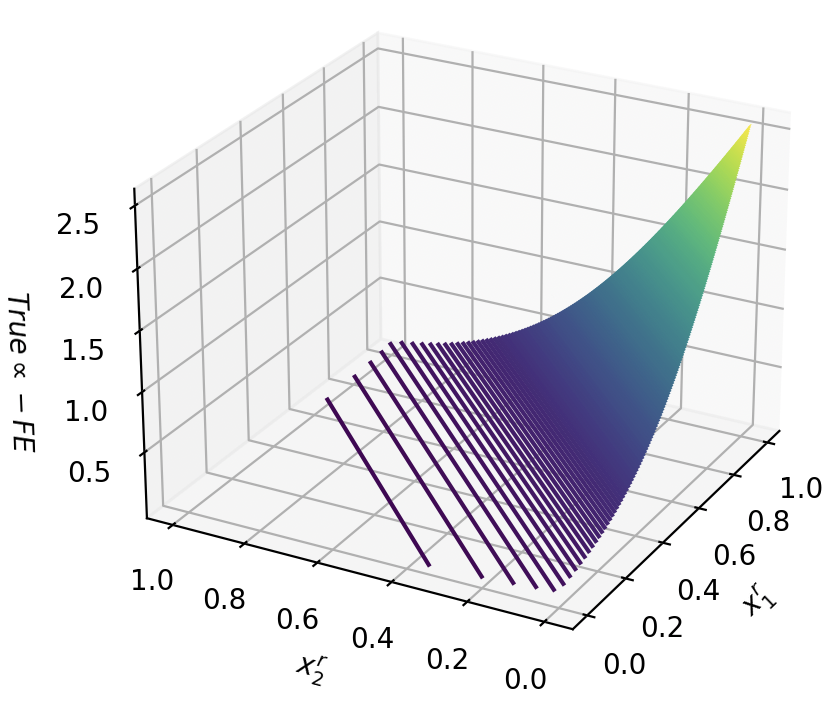}
    \caption{c=5}
    \label{fig:c5}
    \end{subfigure}
    \begin{subfigure}{0.3\textwidth}
    \includegraphics[width=0.9\textwidth]{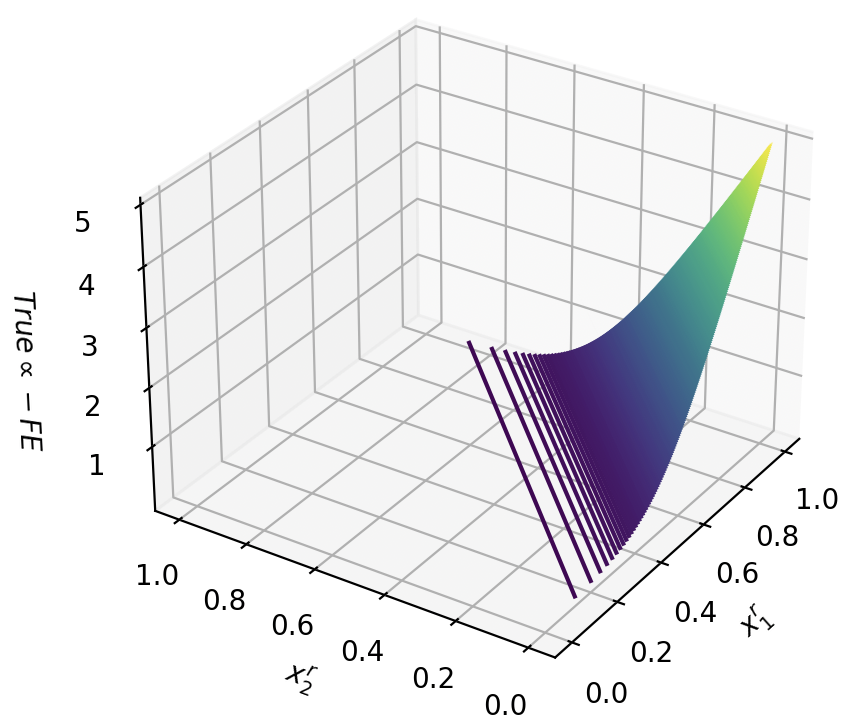}
    \caption{c=10}
    \end{subfigure}
    \vskip -0.3cm
    \caption{Free energy corresponding to an AND gate with different confidence values (Example \ref{orandgates}).}
    \label{fig:and}
\end{figure*}

\begin{figure*}[ht]
    \begin{subfigure}{0.3\textwidth}
    \includegraphics[width=0.9\textwidth]{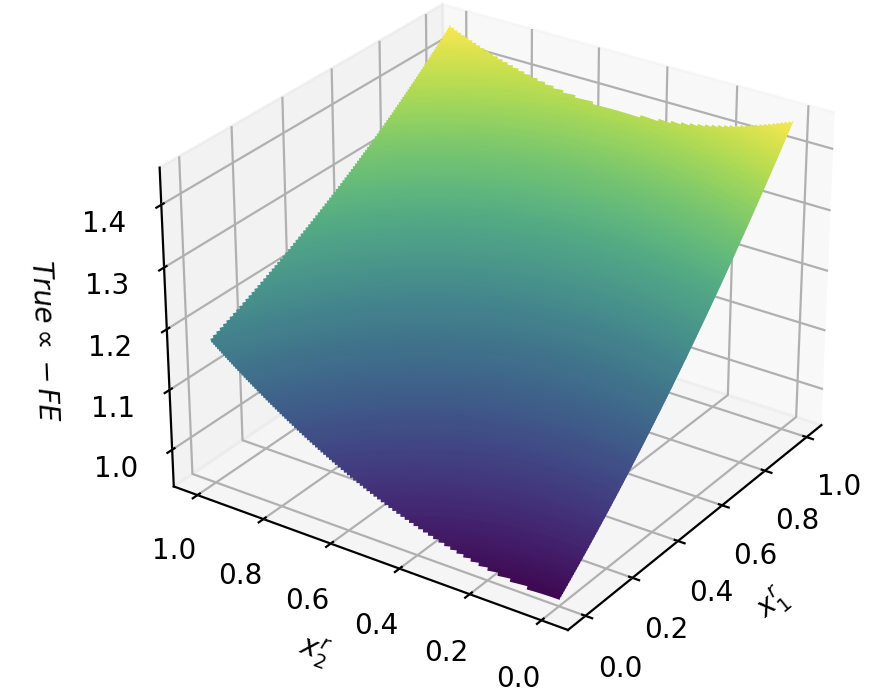}
    \caption{c=1}
    \label{fig:c11}
    \end{subfigure}
    \begin{subfigure}{0.3\textwidth}
    \includegraphics[width=0.9\textwidth]{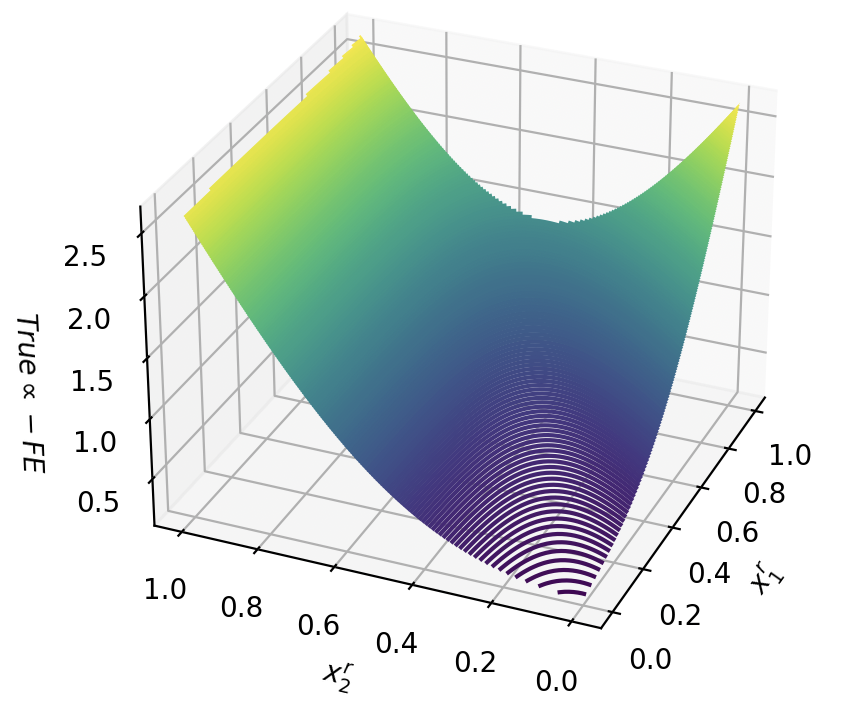}
    \caption{c=5}
    \label{fig:c55}
    \end{subfigure}
    \begin{subfigure}{0.3\textwidth}
    \includegraphics[width=0.9\textwidth]{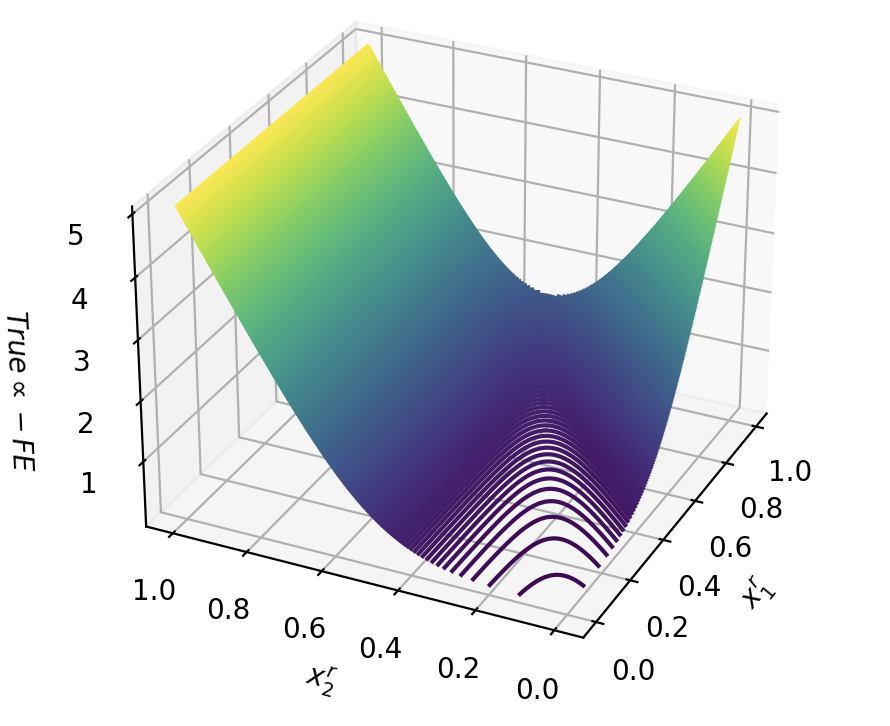}
    \caption{c=10}
    \end{subfigure}
    \caption{Free energy corresponding to an OR gate with different confidence values (Example \ref{orandgates}).}
     \label{fig:or}
\end{figure*}

\begin{example}
\label{orandgates}
An AND-gate $\pr{x}_1 \fzand \neg \pr{x}_2$ is represented by a free energy $FE=-\log(1+\exp(c\times (x_1 - x_2 -0.5)))$. Figure \ref{fig:and} illustrates the correspondence between the free energy and the truth-values for different values of $c$. Similarly, Figure \ref{fig:or} shows the free energy of an OR-gate (that is, a clause) $\pr{x}_1 \fzor \pr{x}_2 $. This clause is transformed into SDNF $(\pr{x}_1 \fzand \neg \pr{x}_2) \fzor \pr{x}_2$ and the corresponding free energy is $FE=-\log(1+\exp(c\times (x_1 - x_2 -0.5))) - \log(1+\exp(c\times(x_2-0.5))$. As expected, the satisfying assignments are those that maximize the negative free energy.
\end{example}

As we have seen already when using LBM as a SAT solver, a conjunctive clause $\varphi_m$ can be represented in an RBM with the energy function $E_m = \sum_j e_j$ and, therefore, the energy function of a CNF will be:
\begin{equation}
\En(\vt{x}) = \sum_mE_m
\end{equation}
The free energy of each clause corresponds to the truth values of the clause, i.e. $\mathbbm{1}(\vt{x}\models\varphi_m) \propto FE_m(\vt{x})$. The free energy of the entire CNF $FE(\vt{x}) = \sum_mFE_m(\vt{x})$, therefore, corresponds to the number of satisfied conjunctive clauses, that is:
\begin{equation}
\begin{aligned}
\sum_m \mathbbm{1}(\vt{x}\models\varphi_m) \propto FE(\vt{x}) \\
\end{aligned}
\end{equation}
An assignment that maximizes the number of satisfying clauses in a MaxSAT problem also minimizes the free energy of the LBM. Consequently, solving MaxSAT problems is equivalent to searching for a state of minimum free energy in the RBM.


\subsection{LBM with Dual Annealing}

When representing CNF in a RBM, one option for solving the MaxSAT problem is to utilize stochastic search methods like Gibbs sampling. However, in our scenario, Gibbs sampling exhibits slow convergence, necessitating potentially extensive computational resources to reach equilibrium. To mitigate this challenge, we have adopted simulated annealing, a meta-heuristic technique renowned for addressing global optimization problems \cite{sa}. Specifically, we employ dual annealing, which seamlessly integrates global and local search strategies to enhance efficiency \cite{XIANG1997216}.

In this approach, the search algorithm initially employs simulated annealing to identify a candidate region within the search space where global optima are likely to be situated. Subsequently, a local search is conducted to precisely pinpoint the location of the global optima. This hybrid strategy combines the exploration capabilities of simulated annealing with the exploitation capabilities of local search, offering a robust and efficient approach to solving MaxSAT problems.

\subsubsection{Experimental Results of LBM for MaxSAT}

As a preliminary evaluation, we compare the performance of LBM at solving MaxSAT problems with Loandra, a state-of-the-art MaxSAT solver. Loandra was chosen as benchmark solver due to its performance in the MaxSAT Evaluations 2023. Experiments were carried out on a set of challenging MaxSAT instances known as MaxCut, each containing 1,260 clauses. Six distinct MaxCut problem instances were chosen from the MaxSAT 2016 benchmark. Both the LBM and Loandra solvers were run on each of the six instances with a timeout limit of 300 seconds for each run. The instances were chosen to represent diverse problem structures so as to evaluate the capabilities of the solvers without making assumptions about the CNF structure. All experiments were conducted on a standard desktop computer with a AMD ryzen7 5800X 8-core processor and 32 GB DDR4 RAM. The main evaluation metric was the number of satisfied clauses, indicating the solver's ability to maximize clause satisfaction within the given time constraints. 


\begin{figure}[ht]
    \centering
    \includegraphics[width=0.9\textwidth]{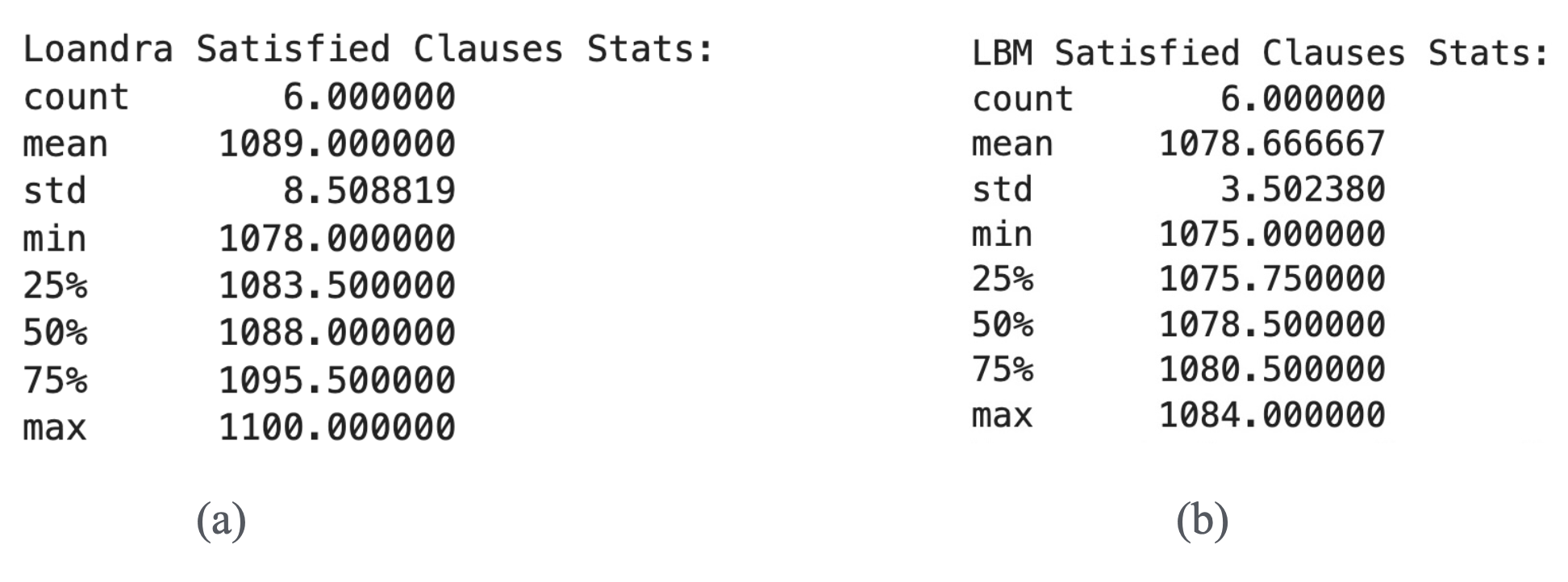}
    \caption{Statistics for clause satisfaction for (a) Loandra (1089 clauses satisfied on average) and (b) LBM for MaxSAT (1078 clauses satisfied on average).}
    \label{fig:statisticsLBM-Loandra}
\end{figure}


\begin{figure}[ht]
    \centering
    \includegraphics[width=0.60\textwidth]{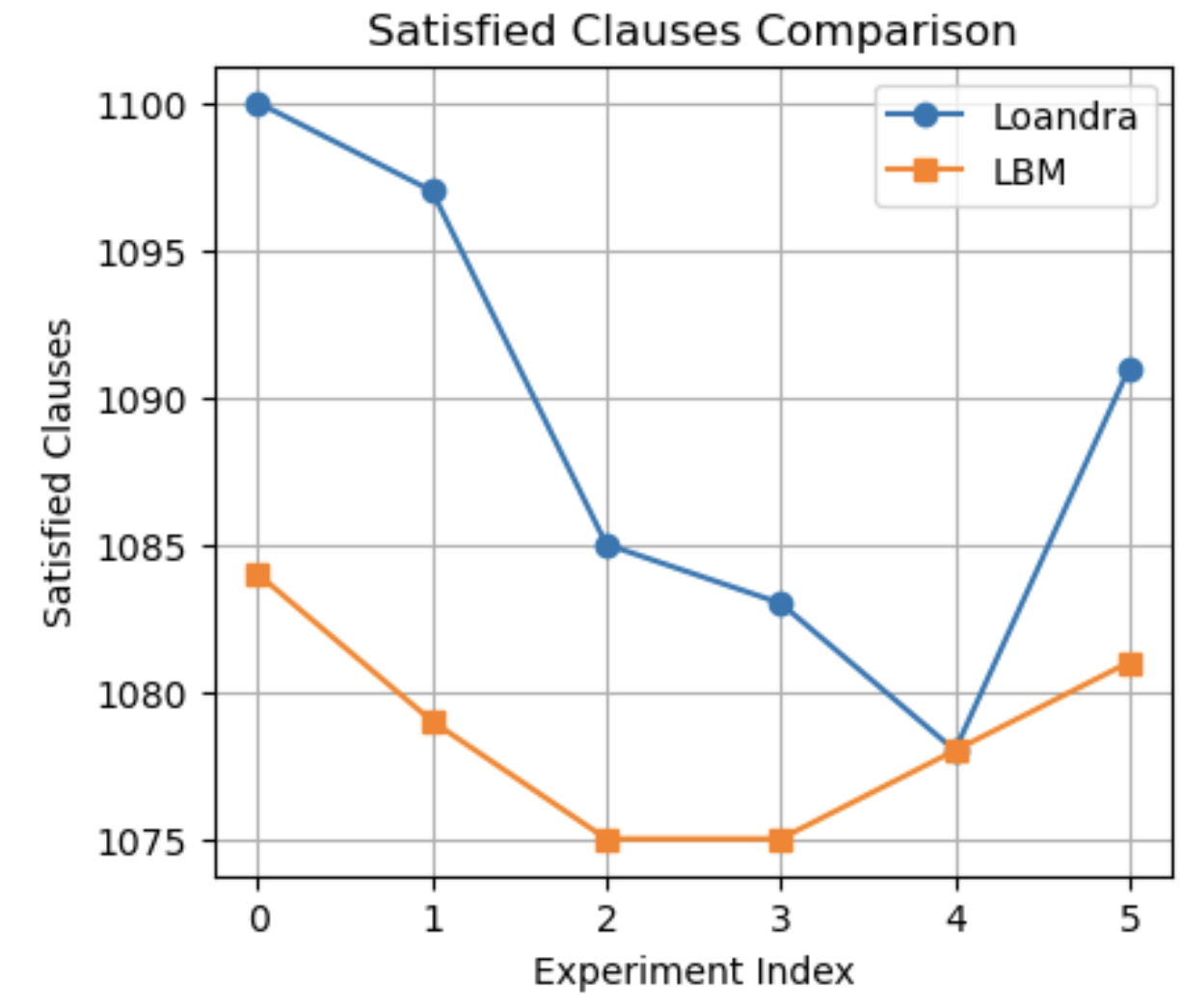}
    \caption{Performance of LBM and Loandra on 6 MaxCut experiments (0 to 5).}
   \label{fig:lbm-loandra-compare}
\end{figure}




Figure~\ref{fig:lbm-loandra-compare} shows the number of clauses satisfied by the symbolic SAT solver Loandra and the LBM for MaxSAT approach in each of the 6 experiments. As a state-of-the-art solver, Loandra performed better than LBM in 5 out of 6 cases. However, the difference is relatively small and further investigation is warranted, with numerous directions for future exploration. One possible direction involves investigating alternative global optimization methods such as evolutionary strategies. Another obvious direction is the implementation of parallel computation to improve scalability of LBM search, similar to the approach employed in RBMSAT. As future work, the task of \textit{Image Sudoku} introduced in \cite{DBLP:journals/corr/abs-2406-09949} is an interesting extension to consider because of the interplay between object recognition and reasoning that is intrinsic to that task and aligned well with LBM's approach integrating learning and reasoning, which we describe next. 


%% file: firstorder.tex
A key development in neurosymbolic AI will be the ability to add verified \textit{modules} to existing networks. An RBM that can be shown to be provably equivalent to a given logical formula could act as one such module. A neurosymbolic module placed on top (at the output) of a larger network may serve to constrain that network's output to satisfy certain properties. In this process, the specification of a neurosymbolic module may benefit from the use of a logical language that is richer (more expressive) than propositional logic. In particular, the use of first-order logic may be required. 

In first-order logic, instead of using propositions and connectives alone, logical \textit{predicates} are used to represent relations among quantified variables. Statements such as $\forall X \exists Y. R(X,Y)$ are used to denote compactly a relation $R$ between variables $X$ and $Y$ in a domain such that for all the values that variable $X$ may admit, there is a value (an instance) of variable $Y$ for which the relation $R$ holds \textit{True}. 

As an example, if we were to rewrite into first-order logic the Boolean logical notation provided earlier for the Sudoku puzzle, we could use a ternary predicate $board(B, P, V)$, in which:
\begin{itemize}
    \item $B$ represents the block index (1 to 4),
    \item $P$ represents the position index within the block (1 to 4), and
    \item $V$ represents the value (1 to 4),
\end{itemize}
to make the problem description a lot more compact.

A first-order representation of the constraints of the Sudoku puzzle would include:
\begin{enumerate}
    \item \textbf{Existence:} $\forall B, P, \exists V. board(B, P, V)$ 
    
    This ensures that every position in every block is filled with at least one value.
    \item \textbf{Uniqueness:}  $\forall B,P, V_1, V_2. (board(B, P, V_1) \land board(B, P, V_2)) \rightarrow V_1 = V_2$ 
    
    This ensures that no position in the board can hold multiple values.

\end{enumerate}

%

We will illustrate one way of integrating LBM as a logical layer on top of deep networks applied to a semantic image interpretation task. The task is to predict the relations between objects and their parts in an image. It requires the use of first-order logic. The knowledge base consists of symbolic facts expressing when an object type is normally part of another object type, e.g. ${part}(\text{Screen},\text{TV})$, where Screen and TV are variables, denoting that TVs have screens. The knowledge base also includes a first-order rule connecting any two visual scenes ($X_1$ and $X_2$) with the symbols of the logic, as follows:

\begin{equation}\label{first-order-rule}
\begin{aligned}
\forall X_1,X_2,\exists T_1,T_2 . (({type}(X_1,T_1) \fzand {type}(X_2,T_2)) \rightarrow \\ ({partOf}(X_1,X_2) \fziff {part}(T_1,T_2)))
\end{aligned}
\end{equation}

\noindent where $X_1$, $X_2$ are real-valued variables representing visual features of objects (an embedding, pixel values, etc.), as done in \cite{Serafini_2016}, and $T_1$, $T_2$ are symbolic variables representing object types. Predicate ${type}$ is $True$ when an object, as defined by its visual features, is deemed to be of a given type (e.g. an object class). Given two visual scenes with their corresponding type classifications, ${type}(X_1,T_1)$ and ${type}(X_2,T_2)$, one visual scene will be part of the other, $partOf(X_1,X_2)$, if and only if the object type of the former is deemed to be part of the object type in the latter, ${part}(T_1,T_2)$.

In order to implement the above rule, we use Faster-RCNN to extract features from object images, from which we build two Neural Network Regressors (NNR)\footnote{To represent first-order logic in LBM, we combine LBM with the Neural Network Regressors. Each NNR represents a predicate in the formulae and outputs a truth-value for that predicate. LBM takes as input the truth-values of the predicates coming from the outputs of the NNRs. In essence, LBM sits on top of the NNRs connecting the predicates according to the connectives of the corresponding logic formulae.}: $\mathcal{N}^{type}$ and $\mathcal{N}^{po}$, as learned functions for ${type}$ and ${partOf}$, respectively, as done in \cite{Ivan_2017}. Finally, we use an autoencoder $\mathcal{N}^{pt}$ to implement the relation ${part}(T_1,T_2)$ between the symbolic variables $T_1$ and $T_2$, following \cite{Tran_2021}. Let $  {p}^  {po}=\mathcal{N}^  {po}(X_1,X_2)$, $  {p}^  {pt}=\mathcal{N}^  {pt}(T_1,T_2)$, $  {p}^  {t_1}=\mathcal{N}^  {type}(X_1,T_1)$, $  {p}^  {t_2}=\mathcal{N}^  {type}(X_2,T_2)$, ${p}^{po}, {p}^ {pt}, {p}^{t_1}, {p}^  {t_2} \in \{0,1\}$, according to some choice of threshold. The first-order rule (\ref{first-order-rule}) can be converted to SDNF, as follows:
{
\begin{align*}
    & (  {p}^  {t_1} \fzand   {p}^  {t_2}) \rightarrow (  {p}^  {po} \fziff  {p}^  {pt})
    \equiv \\ & (  {p}^  {po} \fzand   {p}^  {pt} \fzand   {p}^  {t_1}\fzand   {p}^  {t_2}) \fzor(\neg  {p}^  {po}\fzand \neg  {p}^  {pt}\fzand    {p}^  {t_1}\fzand    {p}^  {t_2})\fzor (\neg   {p}^  {t_1}\fzand   {p}^  {t_2})\fzor \neg   {p}^  {t_2}
\end{align*}
}

\begin{figure}[ht]
    \centering
    \includegraphics[width=0.8\textwidth]{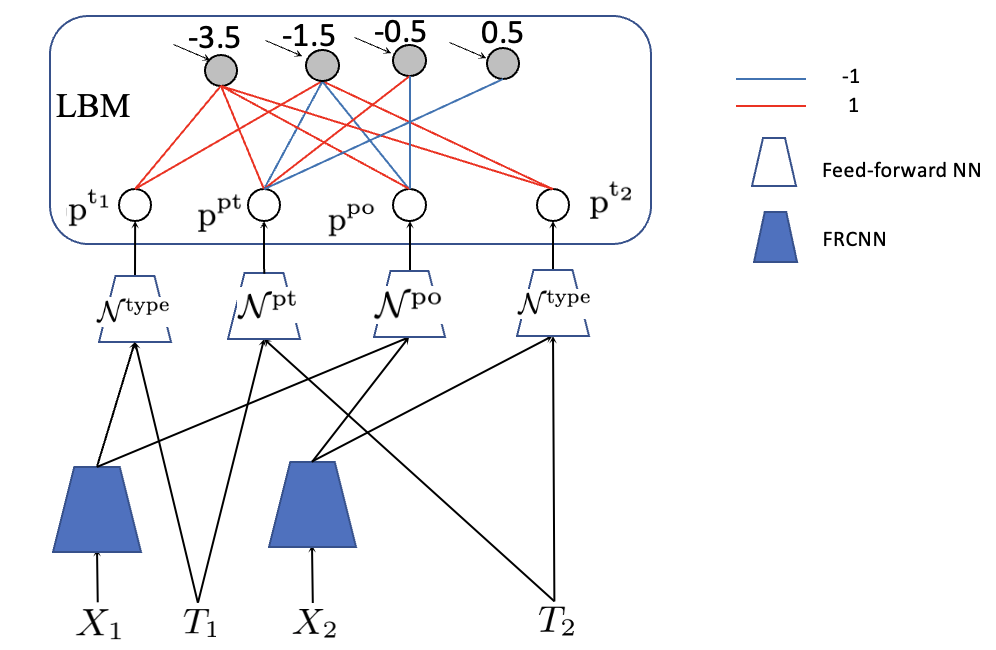}
    \caption{Learning and reasoning about the \textit{PartOf} relation in object images by grounding symbolic concepts into Convolutional Neural Networks and adding a logical layer in the form of a LBM module implementing the rule $({p}^  {t_1} \fzand   {p}^  {t_2}) \rightarrow (  {p}^  {po} \fziff  {p}^  {pt})$.}
    \label{fig:sii}
\end{figure}

From this SDNF, we build a LBM as the \emph{logical layer} on top of the neural networks $\mathcal{N}^{type}$, $\mathcal{N}^{po}$ and $\mathcal{N}^{pt}$. Figure \ref{fig:sii} shows the overall network architecture.


Reasoning in the LBM can inform learning in $\mathcal{N}$ by backpropagating inferred knowledge to update the weights of the CNN, regressor or autoencoder. In particular, we train the entire system by minimizing the following loss function\footnote{$[\mathcal{N}^ {type}(x_1,t_1),\mathcal{N}^ {type}(x_2,t_2)]$ denotes the concatenation of the outputs from the $\mathcal{N}^{type}$ networks; $||x||^2_2$ is the squared Euclidean norm.}:

 \begin{align*}
 &||\mathcal{N}^  {po}(x_1,x_2)-\text{LBM}(  {p}^  {po}|\mathcal{K}(x_1,x_2)||^2_2 +\\ 
 & ||[\mathcal{N}^  {type}(x_1,t_1),\mathcal{N}^  {type}(x_2,t_2)] -\text{LBM}(  {p}^  {t_1},   {p}^  {t_2}|\mathcal{K}(x_1,x_2) )||^2_2
 \end{align*}

\noindent where $x_1,x_2$ and $\mathcal{K}(x_1,x_2)$ are obtained from the training data; $\mathcal{K}$ denotes the knowledge pertaining to $x_1, x_2$, i.e. the type of $x_1$, type of $x_2$, and whether $x_1$ is part of $x_2$. We use $\text{LBM}(  {p}^  {po}|\mathcal{K}(x_1,x_2))$ and $\text{LBM}(  {p}^  {t_1},   {p}^  {t_2}|\mathcal{K}(x_1,x_2)) $ to denote the application of LBM to infer the value of $  {p}^  {po}$ and of the pair $[  {p}^  {t_1},   {p}^  {t_2}]$, respectively. For example, the LBM is used to infer $  {p}^  {po}$, which is used in turn to update $\mathcal{N}^  {po}$. 

Given, for instance, $x_1=\imgobj{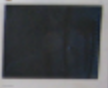}$ and $x_2=\imgobj{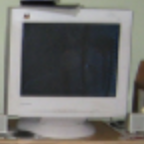}$, let's assume that we do not know whether $x_1$ is part of $x_2$. But, if the $\mathcal{N}^ {type}$ networks tell us that ${type}(x_1,\text{Screen}) \equiv {True}$ and  $ {type}(x_2,\text{TV}) \equiv {True}$ and $\mathcal{K}(x_1,x_2)$ also includes ${part}(\text{Screen},\text{TV})$, the LBM can infer that $  {p}^  {po}$ should be $True$. Finally, this signal from the LBM's reasoning, obtained from the first term of the loss function, is used to update during learning the parameters of $\mathcal{N}^{po}$. Similarly, the second term of the loss function leverages information about the type of objects to update $\mathcal{N}^{type}$.
 
We compared the LBM-enhanced model with three other neurosymbolic systems: Deep Logic Networks (DLN) \cite{Son_2018}, Logic Tensor Networks (LTN) \cite{Ivan_2017, BADREDDINE2022103649}, and Compositional Neural Logic Programming (CNLP) \cite{Tran_2021}. The task and data set used were the same as in \cite{Ivan_2017}, with the exception of the rule $({p}^  {t_1} \fzand   {p}^  {t_2}) \rightarrow (  {p}^  {po} \fziff  {p}^  {pt})$ only used by LBM. The \textit{area under the curve} (AUC) results shown in Table \ref{tab:sii_result} indicate that the use of this single first-order rule for reasoning with the LBM and training of the Faster-RCNN model, also used in \cite{Ivan_2017}, produces a higher performance than LTN in the prediction of the part-of relation in images. LBM's performance is comparable to that of CNLP. For the object type prediction, the LBM model performs better than CNLP, DLN and LTN.
 
\begin{table}
\centering
\caption{Comparison of neurosymbolic approaches; area under the curve (AUC) in the semantic image interpretation task. Source: \cite{Tran_Garcez_2023}.
    \label{tab:sii_result}}
{%
\begin{tabular}{@{}ccc@{}}
\toprule
& Object type (AUC) & Part-of (AUC)  \\
\midrule
DLN & $0.791 \pm 0.032$ & $0.605 \pm 0.024$ \\
CNLP & $0.816 \pm 0.004$ & $\mathbf{0.644\pm 0.015}$ \\
LTN& $0.800$ & $0.598$\\
LBM & $\mathbf{0.828 \pm 0.002}$ & $\mathbf{0.645 \pm 0.027}$\\ 
\toprule
\end{tabular}}
\end{table}

%% file: artur2.tex
\label{sec:discussion}

We introduced an approach and neurosymbolic system to reason about symbolic knowledge in an energy-based neural network. Differently from most LLM approaches and CoT, where reasoning capabilities are expected to emerge and are measured post-hoc using benchmarks, we use logic to provide a formal definition of reasoning. Only once a well-defined semantics is provided, one can show correspondence between networks and various forms of reasoning. We showed equivalence between propositional logic and RBMs. The finding led to a novel system, named Logical Boltzmann Machines integrating learning and reasoning in neural networks. Future work will focus on scaling up the application to SAT and learning from data and knowledge. Extensions include applications of weighted clauses, weighted SAT with parallel implementation as well as evaluations on relational learning tasks.

Equipped with a proof and algorithm showing how RBMs can implement a category of required constraints, it is possible to imagine how an RBM may be added as a \textit{module} to an existing network imposing such constraints on the network. This RBM module becomes a \textit{verifiable component} of the system, implementing for example a fairness or safety requirement as argued for in \cite{eleonora}. Next, we discuss how this simple idea may open up directions for research addressing some of the biggest challenges for current AI: data efficiency, fairness, safety and ultimately trust.




\subsection{Nonmonotonic Logic}

Consider the task of commonsense reasoning, the human-like ability to make sense of ordinary situations, such as making judgments about the nature of objects. It encompasses intuitive psychology (e.g. reasoning about intentions) and naive physics (a natural understanding of the physical world, including spatial and temporal reasoning). Commonsense reasoning requires an ability to jump to conclusions based on incomplete information, and to retract or revise such conclusions when more information become available. There have been many attempts to formalize commonsense, not least the work of John McCarthy who was responsible for coining the term Artificial Intelligence. Because it requires jumping to conclusions and therefore handling logical inconsistencies in a non-classical way, commonsense reasoning is modeled by non-monotonic logics, some of which are undecidable. LLMs have achieved much better results than logical formalizations of commonsense on reasoning benchmarks. It turns out that it is easier to learn commonsense from data than to formalize it logically. Yet, making sense of what has been learned by LLMs has been proven to be a daunting task. Making sense of what has been learned might help tremendously with the efforts to understand the limits of the formalization of commonsense in logic. As neural networks start to be deployed successfully in various fields of scientific discovery, achieving a true understanding of the processes at play will demand such an ability to explain the network’s reasoning. 

\subsection{Planning}

AGI will require the ability to plan towards a goal and the ability to ask questions in order to achieve that goal. Planning requires the ability to break-down goals into sub-goals while reasoning with partial information over time. Having a goal, in turn, requires the provision of a description of the current state and the goal state, and a mechanism that reduces the difference (e.g. some distance function) between the current state and the goal state by changing the current state in a reasonable way. If the mechanism isn’t working, one may decide to change the goal or even change the mechanism itself. 

Whether in Google's AlphaGo or Sudoku, the goal state is to win the game, and because it is a closed environment, simulation can be used to learn to minimize the distance function without the need for an explicit description. In open-ended situations, the problem becomes much harder. An explicit description is one that can be manipulated by asking questions: “what might happen if I were to make this or that change?” without making the change. An explicit description needs, therefore, to be amenable to symbolic manipulation. We argue that in open-ended scenarios, an explicit description needs furthermore to be abstracted from the situation given only a few examples. Reasoning in the form of symbol manipulation on that abstract description can then take place that will be of a different nature from reasoning carried out in terms of pattern matching with similarity and distance functions. 

\subsection{Learning from its Mistakes}

An AGI system should be able to learn from its mistakes, interact with users by asking questions, describing its understanding and improving its performance in a controlled way towards a goal, even if changing the goal and the mechanism for performance improvement. Controlled at the level of its symbolic description, the system can be made safe. With reasoning taking place at both the pattern matching and the more abstract levels, as advocated by D. Kahneman’s Thinking: Fast and Slow \cite{kahneman2011thinking}, the system will be able to adapt to novelty from only a few examples, check its understanding, multi-task and reuse knowledge from one task to another thus improving data and energy efficiency in comparison with the demands of current AI systems.

Adapting to novelty (therefore solving the long-tail distribution problem and out-of-distribution problem in Machine Learning) requires creating compact representations (in the brain or the mind) but also being able to change that representation from time to time in order to obtain new insight. It is the change of representation that allows one to look at a problem from a new angle to obtain new insight. Think of the difference between two computer programs, both correct for their purpose, one so-called \textit{spaghetti} code and the other an example of what programmers like to refer to as \textit{beautiful} code. The former may be faster to run, but the latter needs to be neat, easy to understand and useful to update and reuse. This is the core challenge of the latest research in neurosymbolic AI: extraction of relevant descriptions at the right level of abstraction from complex neural networks, sound application of reasoning and learning with various forms of representation - spatial, temporal, epistemic, normative, multimodal, nonmonotonic - and efficient knowledge and data reuse and extrapolation to multiple tasks in different application domains. 

Consider the kind of program learned by the latest GPT-based chatbots. Transformer neural networks work, in essence, by mapping sets to sets and not sequences to sequences (GPT-based chatbots require positional encoding to handle sequences). Interestingly, in neurosymbolic AI, the computation of the semantics of various logics by neural networks is also done by mapping sets to sets recursively \cite{OdenseAI}. A great innovation of deep learning was multi-headed attention which is similar to representing binary relations in neurosymbolic networks. In certain domains of application, though, such as protein interaction, having the ability to represent not just binary but n-ary relations such as e.g. \textit{bond(Compound\_Name, Bond\_Type, Atom\_1, Atom\_2)} may be very useful. This is the case in the Mutagenesis task and data set, where a chemical compound may have a bond of a certain type between any two atoms and the goal is to identify mutagenic compounds given the atom bonds. Hypergraph neural networks have been used recently to represent and learn n-ary relations \cite{Hypergraph}. Finally, the neurosymbolic (NeSy) framework of fibring neural networks \cite{fibring} has been shown recently to offer a common representational foundation for both graph neural networks and transformers via a proof of correspondence with propositional modal logic with self-fibring. These are some of the exciting recent developments in neurosymbolic AI (see the NeSy conference series for more \cite{NeSy2024}) contributing to both a better understanding of deep learning and the development of new formalisms for learning and reasoning. 

%% file: artur3.tex
Many influential leaders have been pointing out the risks of current AI and arguing for the adoption of regulation. While it is clear that worldwide regulation is not achievable in the current geopolitical climate (see \cite{chip}), an alternative argument is that digital technology itself can offer, as part of an adequate accountability ecosystem, a new path to safer AI. In this new path, neural models can be validated symbolically by adopting the neurosymbolic cycle: train a little, reason a little, repeat. This is quite different from the current \textit{scale-is-all-you-need} approach or what the EU AI Act has achieved. Regulation without accountability tends to increase weak competitiveness and may not decrease risks.

At first impression, the need for accountability in AI and the risks of current AI may seem to be quite disconnected from the technical contributions of this paper. However, we argue that the kind of formalization offered here is key to accountability, fairness and ultimately a safer AI. If neurosymbolic AI can show that compact network modules behave according to a given formal semantics then these modules can be composed in ways that will offer guarantees to the overall system. Of course, this continues to be an important research challenge, but results such as the ones reported in this paper point to an alternative to the current approach to AI, best illustrated by Figure \ref{fig:sii}, where a requirement (or a guardrail) can be implemented as a neurosymbolic network module on top of an existing complex network. For this reason, we conclude with a short summary and opinion on the need for accountability in AI.

The need for accountability in AI is now center stage, as indicated by the following quote from \cite{openAIchaos}: “A long-standing concern among analysts of AI development is the possibility of a race to the bottom in which multiple players feel pressure to neglect safety and security challenges in order to remain competitive. Perceptions - and therefore signals - are key variables in this scenario. Most actors would presumably prefer to have time to ensure their AI systems are reliable, but the desire to be first, the pressure to go to market, and the idea that competitors might be cutting corners can all push developers to be less cautious. Accordingly, signaling has an important role to play in mitigating race-to-the-bottom dynamics. Parties developing AI systems could emphasize their commitment to restraint, their focus on developing safe and trustworthy systems, or both. Ideally, credible signals on these points can reassure other parties that all sides are taking due care, mitigating pressure to race to the bottom”. In \cite{AccountAI}, the authors go further, arguing for an accountability in AI ecosystem. They propose to map out the general principles of AI into industry-specific mechanisms, having stated as early as 2021: “at present the ecosystem is unbalanced, which can be seen in the failures of certain mechanisms that have been attempted by leading technology companies. By taking an ecosystem perspective, we can identify certain elements that need developing and bolstering in order for the system as a whole to function effectively. Corporate governance mechanisms such as standardized processes and internal audit frameworks, leading up to potential external accreditation, need to be made to work together in ways that go beyond regulatory requirements, especially in technologies’ early period of evolution and deployment when regulation lags practice.”

As part of a case study carried out with a global software provider operating primarily in the gambling sector with a focus on online gambling, \cite{AccountAI} reports on the use of AI to help reduce harm from gambling. The application of AI in responsible gambling has been a relevant use case because of the high regulatory focus, divergent regulatory perspectives worldwide, and a longstanding debate over ethical dilemmas relating to an increase in gambling addiction. Results are drawn from the risk profiling of gambling behavior using neural networks and explainability. The neural network performs classification of problem gambling. Explainable AI evaluates indirect gender bias and the need for algorithmic fairness. Results are analyzed in connection with the proposed accountability ecosystem and its operationalization. The AI accountability ecosystem has as stakeholders: corporate actors, market counterparts, civil society and government, alongside mechanisms such as internal auditing, external accreditation, investigative journalism, risk-based regulation and market shaping. Two key elements of the accountability ecosystem are discussed in detail: (i) interventions to reduce bias and (ii) increased transparency via model explainability. The benefits of having an industry-specific accountability process are illustrated in that it can be documented, reviewed, benchmarked, challenged and improved upon, “both to build trust that the underlying ethical principle is being taken seriously and to identify specific areas to do more.” \cite{AccountAI}. The paper’s conclusions support the importance of industry-specific approaches in the operationalization of accountability principles in AI, noting how different metrics, priorities and accountability processes arise in online gambling compared to what might arise in other industries. Taken alongside relevant regulatory efforts on information security and privacy, accountability in AI is expected to reduce the risks of imbalances in regulation. 


Widespread use of GPT-style chatbots is expected to increase productivity but also magnify errors, as humans become complacent in the use of the technology. When trying to distinguish genuine from malicious websites, people have learned over the years to look for grammatical errors, the quality of images and other cues. Learning whether or not to trust the output of LLMs is much harder. At this unique juncture when AI leaves the research laboratory and enters everyday life, new ways of doing the things that we are used to and take for granted will need to be decided upon and implemented quickly, until a better way of doing AI comes that will offer safety guarantees to AI systems.